\documentclass[iicol,pdflatex,sn-mathphys-num]{sn-jnl}

\usepackage{graphicx}%
\usepackage{multirow}%
\usepackage{amsmath,amssymb,amsfonts}%
\usepackage{amsthm}%
\usepackage{mathrsfs}%
\usepackage[title]{appendix}%
\usepackage{xcolor}%
\usepackage{textcomp}%
\usepackage{manyfoot}%
\usepackage{booktabs}%
\usepackage{algorithm}%
\usepackage{algorithmicx}%
\usepackage{algpseudocode}%
\usepackage{listings}%

\theoremstyle{thmstyleone}%
\theoremstyle{thmstyletwo}%

\theoremstyle{thmstylethree}%

\begin{document}

\title[Article Title]{Segmentation of Ink and Parchment in Dead Sea Scroll Fragments}

\author*[1]{\fnm{Berat} \sur{Kurar-Barakat}}\email{berat@tauex.tau.ac.il}

\author[1]{\fnm{Nachum} \sur{Dershowitz}}\email{nachum@tau.ac.il}

\affil*[1]{\orgdiv{Blavatnik School of Computer Science}, \orgname{Tel Aviv University}, \city{Tel Aviv}, \country{Israel}}

\abstract{
The discovery of the Dead Sea Scrolls over 60 years ago is widely regarded as one of the greatest archaeological breakthroughs in modern history. Recent study of the scrolls presents ongoing computational challenges, including determining the provenance of fragments, clustering fragments based on their degree of similarity, and pairing fragments that  originate from the same manuscript---all tasks that require focusing on individual letter and fragment shapes. This paper presents a computational method for segmenting ink and parchment regions in  multispectral images of Dead Sea Scroll fragments. Using the newly developed Qumran Segmentation Dataset (QSD) consisting of 20 fragments, we apply multispectral thresholding to isolate ink and parchment regions based on their unique spectral signatures. To refine  segmentation accuracy, we introduce an energy minimization technique that leverages  ink contours, which are more distinguishable from the background and less noisy than  inner ink regions. Experimental results demonstrate that this Multispectral Thresholding and Energy Minimization (MTEM) method achieves significant improvements over traditional binarization approaches like Otsu and Sauvola in parchment segmentation and is successful at delineating ink borders, in distinction from holes and background regions.
}


\maketitle

\section{Introduction}

The discovery of the Dead Sea Scrolls (DSS) over 60 years ago is one of the greatest archaeological breakthroughs in modern history. These are ancient texts, written or copied mainly between the 2nd century BCE and the 2nd century CE. To ensure their preservation and accessibility, the Israel Antiquities Authority (IAA) has digitized the fragments using multispectral high-resolution imaging (Figure~\ref{fig:ms_images}). 

\begin{figure}[ht]
  \centering
  \includegraphics[width=.741\linewidth]{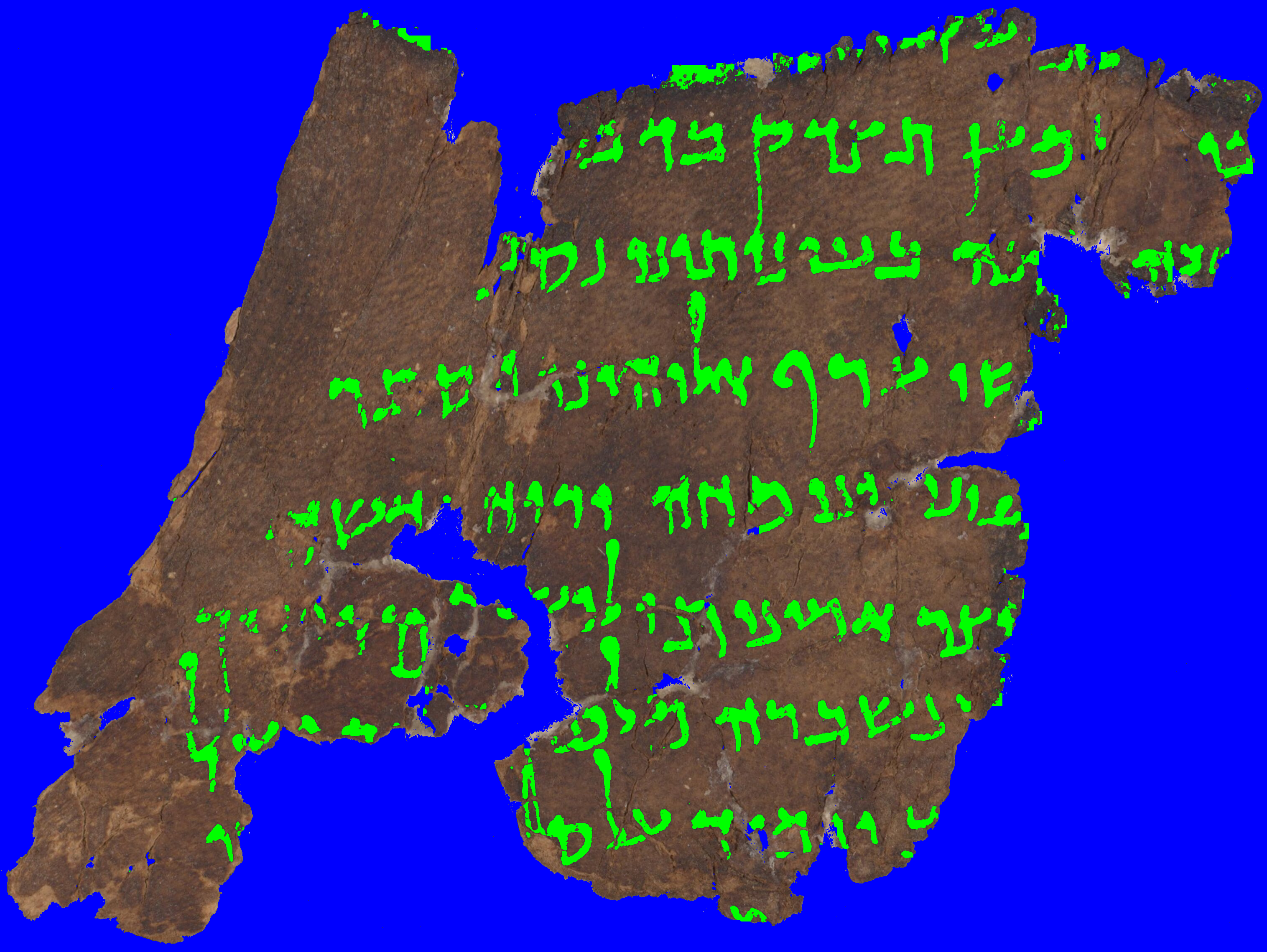}
  \caption{Ink and parchment segmentation using MTEM}
  \label{fig:124_001_greenink_blueback}
\end{figure}

Computational analysis of the DSS often requires binarized letters or binarized fragments (Figure~\ref{fig:124_001_greenink_blueback}). 
There are several challenges in binarizing the DSS images. Similar to many other historical manuscripts, the DSS collection suffers from document degradation problems. Due to aging, the ink often shows fading effects, and the parchment shows darkening effects~\cite{maor2020parchment}. Additionally, black was chosen as the background for the photographs because it does not reflect light, which could have affected the image and reduced the system's stability. However, this choice leads to a most severe issue: the low contrast between ink, background, and holes.

Otsu's method~\cite{4310076} is one of the most commonly used thresholding techniques. This unsupervised and non-parametric method automatically selects a global threshold based on the grayscale histogram of a given image without prior information. The Otsu method performs well when the image has a uniform background; unfortunately, most historical manuscripts do not fall into this category. This can be addressed using small local patches of the target image through local adaptive thresholding methods such as Sauvola's algorithm~\cite{sauvola1997adaptive}. While this performs better than global thresholding, it often shows poor performance in discriminating among multiple classes.

Other approaches to DSS binarization leverage multispectral imaging to exploit learning models. Lavee~\cite{lavee2013computer} trains an SVM classifier on grayscale values across multiple wavelengths to classify each pixel as foreground or background, using baseline labeling from Sauvola binarization. Zarai et al.~\cite{zarai2013integrating} further enhances this approach by incorporating images of old fragments from the 1950s, which were photographed on a white background, into the SVM training while still using Sauvola binarization as the baseline labeling.

Dhali et al.~\cite{dhali2019binet} designed BiNet, an end-to-end binarization approach for Dead Sea Scroll fragments based on the U-Net architecture \cite{ronneberger2015u}. This method aims to segment only the ink, defining the ink as the foreground class and all other materials as the background class. While their approach significantly outperforms unsupervised methods, a major challenge is that the labeling task requires approximately 4--8 hours of a paleographic expert's time per fragment.

Due to the high cost of labeling, we opted to develop a method independent of manual labeling. We take advantage of the multispectral imaging provided by the IAA. Each fragment was photographed in 12 different wavelengths, seven in the visible light spectrum and five in the near-infrared. We first analyzed the pixel value trends of different regions across the 12 bands (Figure~\ref{fig:five_regions_ms_trends}). Based on these trends, we used the 12th and 1st band values and their differences to threshold and identify approximate regions for parchment, background, rice paper, holes, and ink contours. We then applied energy minimization to refine the ink contours by stretching noisy ink contours between parchment and all other regions: rice paper, holes, and background. Subsequently, we used energy minimization to determine the ink regions by stretching the inverse parchment between clean ink contours and parchment. Finally, we combined the ink regions and parchment regions to achieve parchment segmentation (Figure~\ref{fig:method_pipeline}).

Multispectral imaging is often used to enhance readability in degraded historical documents, as shown in studies of ancient Hebrew ostraca, where near-infrared wavelengths improved contrast between ink and background surfaces~\cite{faigenbaum2012multispectral}. In contrast, our approach for the DSS focuses on precise ink segmentation, grouping pixels that belong to ink regions for accurate classification, rather than solely increasing legibility.

Our method is comprehensive, exhaustively segmenting each pixel in a DSS fragment image and assigning it to one of the ink, parchment, or background classes. It does not rely on morphological or edge detection operations and does not make assumptions about component size. Thus, the segmentation of a part of an image is consistent with its segmentation as part of the entire image. To support benchmarking of segmentation of ink and parchment regions in Dead Sea Scrolls fragment images, we have created the Qumran Segmentation Dataset (QSD), which is publicly available at \url{https://www.cs.tau.ac.il/~berat}.

\section{Qumran Segmentation Dataset (QSD)}

The Qumran Segmentation Dataset (QSD) is designed to benchmark studies targeting the segmentation of ink and parchment regions in Dead Sea Scrolls fragment images. Segmenting these regions enables a more detailed investigation of the scrolls' textual content and material properties, laying the foundation for further computational analysis and scholarly research. This section outlines the digitization process of the scrolls, describes the dataset's composition, the preprocessing applied to prepare the data, and details the data annotation process.

\subsection{Digitization of the Dead Sea Scrolls}

The IAA initiated a digitization project in 2011 aimed at preserving the Dead Sea Scrolls and increasing access to them. 
Full-color and infrared images are publicly available online through the Leon Levy Dead Sea Scrolls Digital Library \cite{leon}.
This allows scholars and the general public to explore them, while at the same time addressing deterioration caused by environmental factors and improper handling. 

\begin{figure*}[ht]
  \centering
  \includegraphics[width=.95\linewidth]{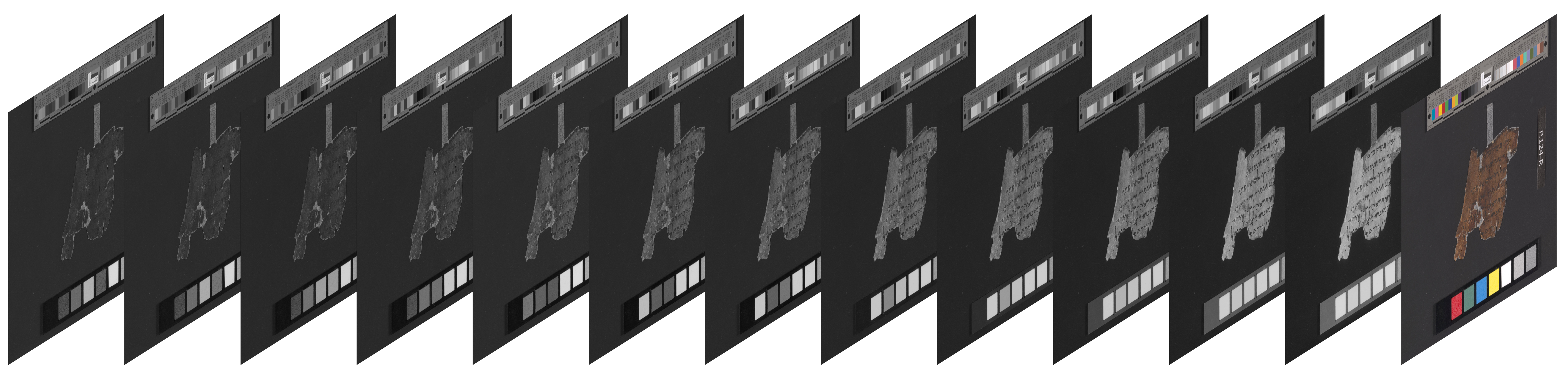}
  \caption{Multispectral imaging of a DSS fragment at different wavelengths. The images represent bands 1 to 12 from left to right, followed by the color image on the far right. The first seven images are in the visible light spectrum, and the remaining five are in near-infrared.}
  \label{fig:ms_images}
\end{figure*}

During  digitization, each fragment is stabilized using rice tissue paper to limit movement and is photographed in 12  wavelengths (Figure~\ref{fig:ms_images}), as listed in Table~\ref{tab:wavelengths}.

\begin{table}[ht]
\caption{Multispectral wavelengths used for photographing each fragment, comprising seven wavelengths in the visible light spectrum and five in the near-infrared spectrum.}
\label{tab:wavelengths}
\begin{tabular}{@{}ll@{}}
\toprule
Visible light wavelengths  & Near-infrared wavelengths \\
\midrule
445nm -- Royal Blue            & IR706nm \\
475nm -- Long Blue             & IR728nm \\
499nm -- Cyan                  & IR772nm \\
540nm -- Green                 & IR858nm \\
595nm -- Amber                 & IR924nm \\
638nm -- Red                   &        \\
656nm -- Deep Red              &        \\
\botrule
\end{tabular}
\end{table}

\begin{figure}[htb]
  \centering
  \includegraphics[width=.95\linewidth]{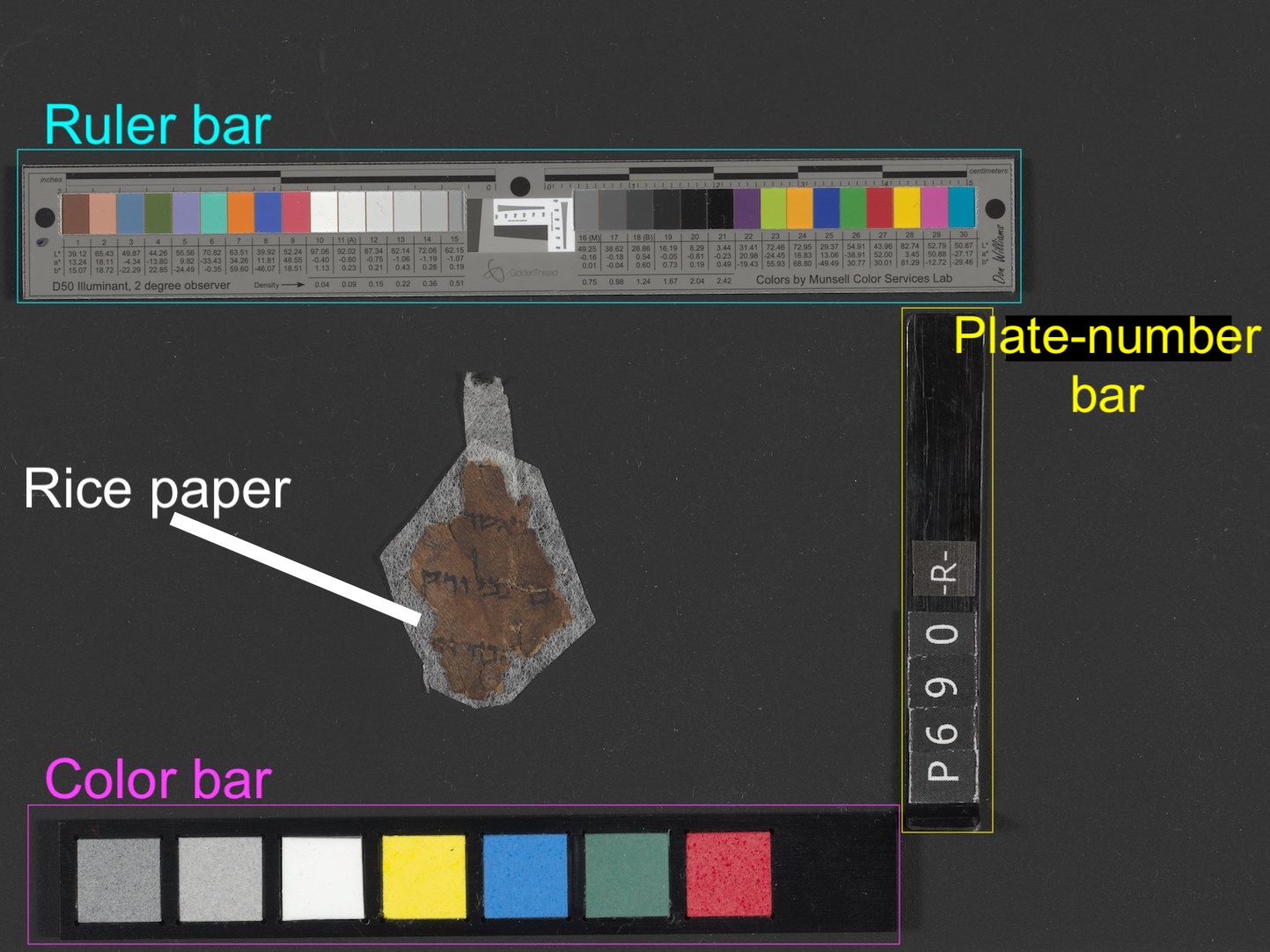}
  \caption{An example IAA full-color image showing the color bar, ruler bar, plate-number bar, and the rice paper used to stabilize the fragment.}
  \label{fig:iaa_bars}
\end{figure}

All of the visible light wavelengths are then combined to create a full-color digital image (Figure~\ref{fig:iaa_bars}). All  images have a resolution of $7216\times 5412$ pixels, regardless of the actual size of the fragments. In the digital library, the full-color image and the last band image are accessible to everyone, while the other exposures are available to researchers upon request.

Although a blue background would have made it easier to computationally isolate fragments from their background by providing high contrast with the parchment and ink, black was chosen to minimize light reflection \cite{Oprescu2018}. A two-part water-based urethane coating \cite{avian} with low reflectance over a wide wavelength range was applied to a stone surface to prevent background light from affecting the captured images. IAA images typically contain three types of bars: a color bar, a ruler bar, and a plate-number bar. The positions of these bars are identical in all band images of a given fragment but vary across different fragments and this variation causes challenges in detecting the fragment in the image (Figure~\ref{fig:iaa_bars}).

\subsection{Dataset composition and preprocessing}

\begin{figure*}[tb]
    \centering
    \begin{minipage}{0.24\textwidth}
        \centering
        \includegraphics[width=\textwidth]{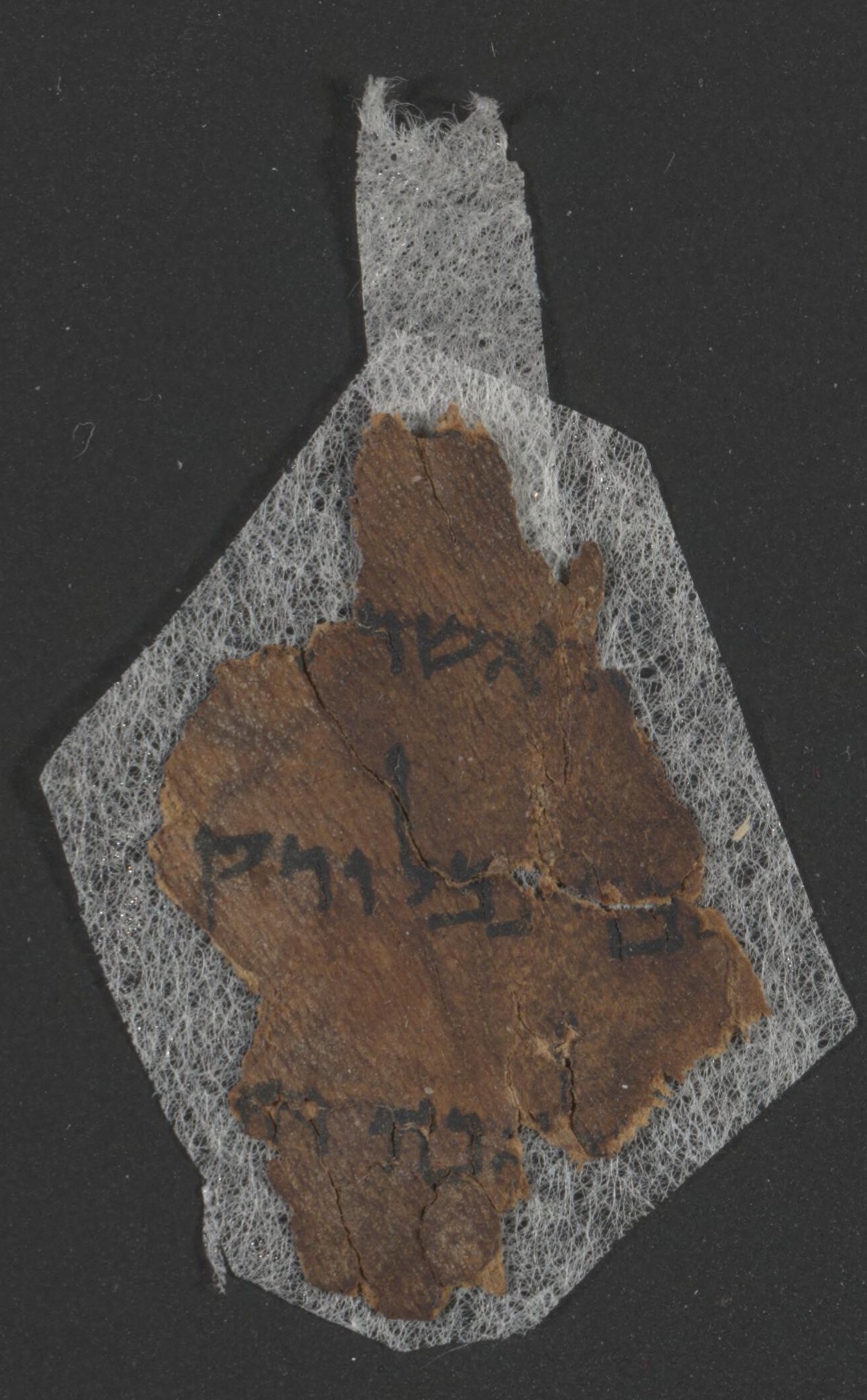}
        \vspace{2mm}
        \text{Full-color}
    \end{minipage}
    \begin{minipage}{0.24\textwidth}
        \centering
        \includegraphics[width=\textwidth]{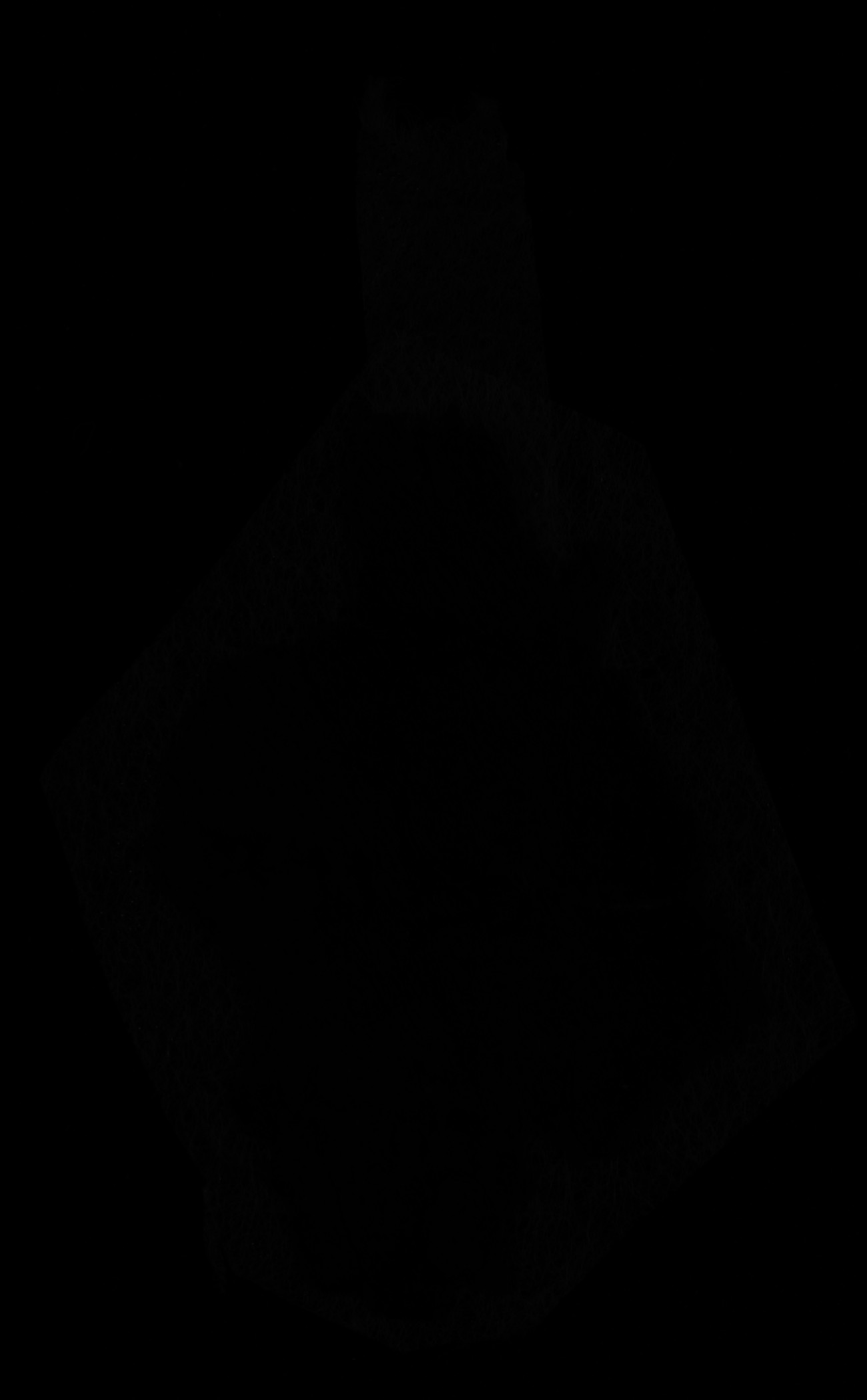}
        \vspace{2mm}
        \text{First-band}
    \end{minipage}
    \begin{minipage}{0.24\textwidth}
        \centering
        \includegraphics[width=\textwidth]{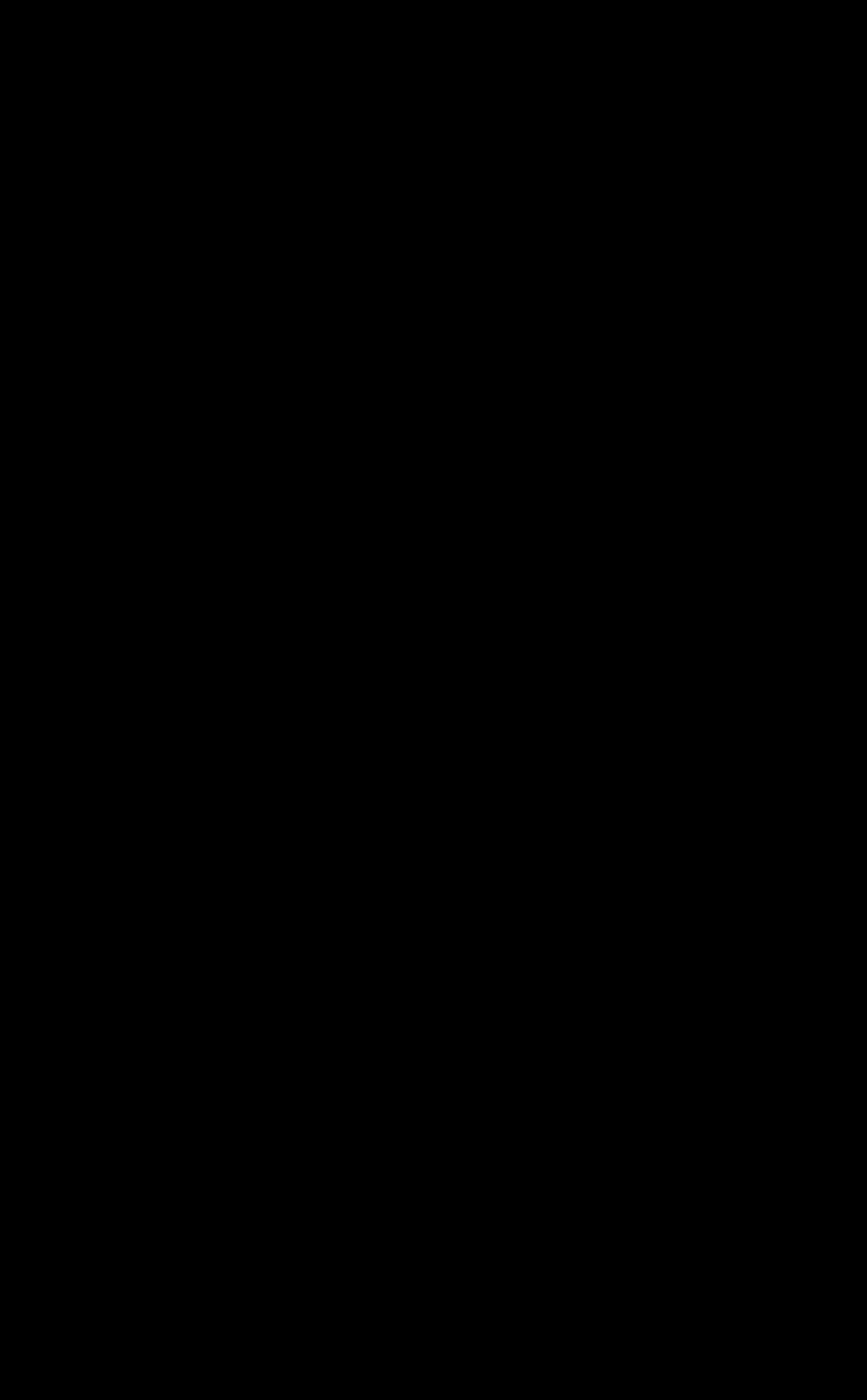}
        \vspace{2mm}
        \text{Last-band}
    \end{minipage}
    \begin{minipage}{0.24\textwidth}
        \centering
        \includegraphics[width=\textwidth]{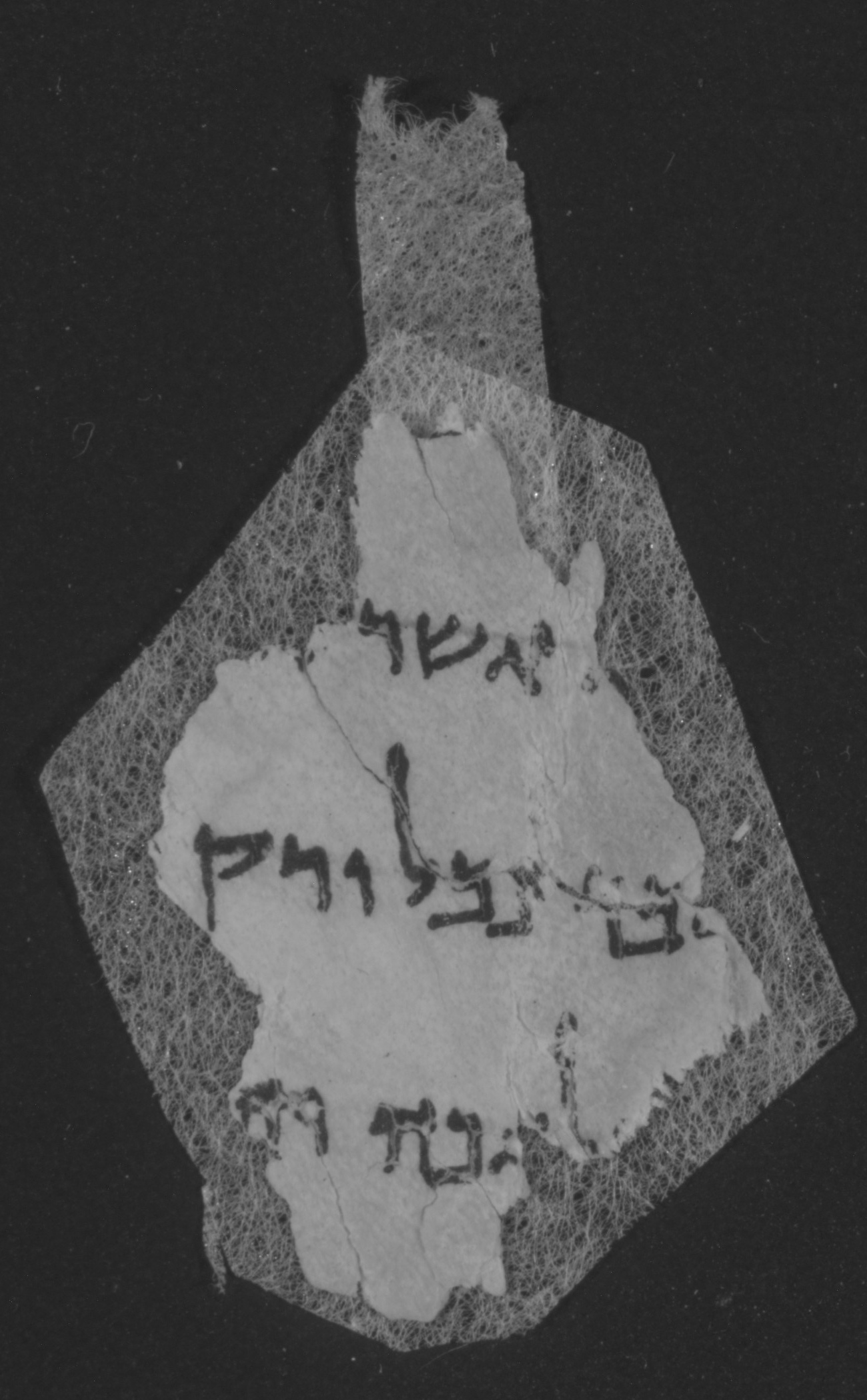}
        \vspace{2mm}
        \text{Normalized last-band}
    \end{minipage}
    \caption{The QSD dataset includes full-color, first-band, last-band, and normalized last-band images of each fragment, cropped to exclude the color bar, ruler bar, and plate-number bar. The first-band and last-band images appear dark due to the narrow range of pixel values in the 16-bit format. The full-color image and the normalized last-band image are useful for qualitative evaluation.}
    \label{fig:dataset_images}
\end{figure*}

The QSD consists of images for 20 randomly selected fragments. For each fragment, the dataset includes a full-color image, a first-band image, a last-band image from the multispectral imaging, and a normalized last-band image (Figure~\ref{fig:dataset_images}). Each image was processed by cropping the rectangular region containing the fragment and rice paper while excluding the color bar, ruler bar, and plate-number bar \cite{brown2024segmenting}. This preprocessing reduces the dataset size, as a single full-color image exceeds 5 MB, and a single band image exceeds 70 MB. This makes the dataset more manageable and ensures that the focus remains on segmentation instead of detection.

The full-color images are 8-bit, have three channels, and are saved in JPEG format, with pixel values ranging from 0 to 255 for each channel. The normalized last-band images are also 8-bit, single-channel, and saved in JPEG format, with pixel values ranging from 0 to 255. In contrast, the multispectral first-band and last-band images are 16-bit, single-channel TIFF files, with pixel values ranging from 0 to 65535.

\subsubsection{Characteristics of first-band and last-band images}

\begin{figure}[htb]
  \centering
  \includegraphics[width=.95\linewidth]{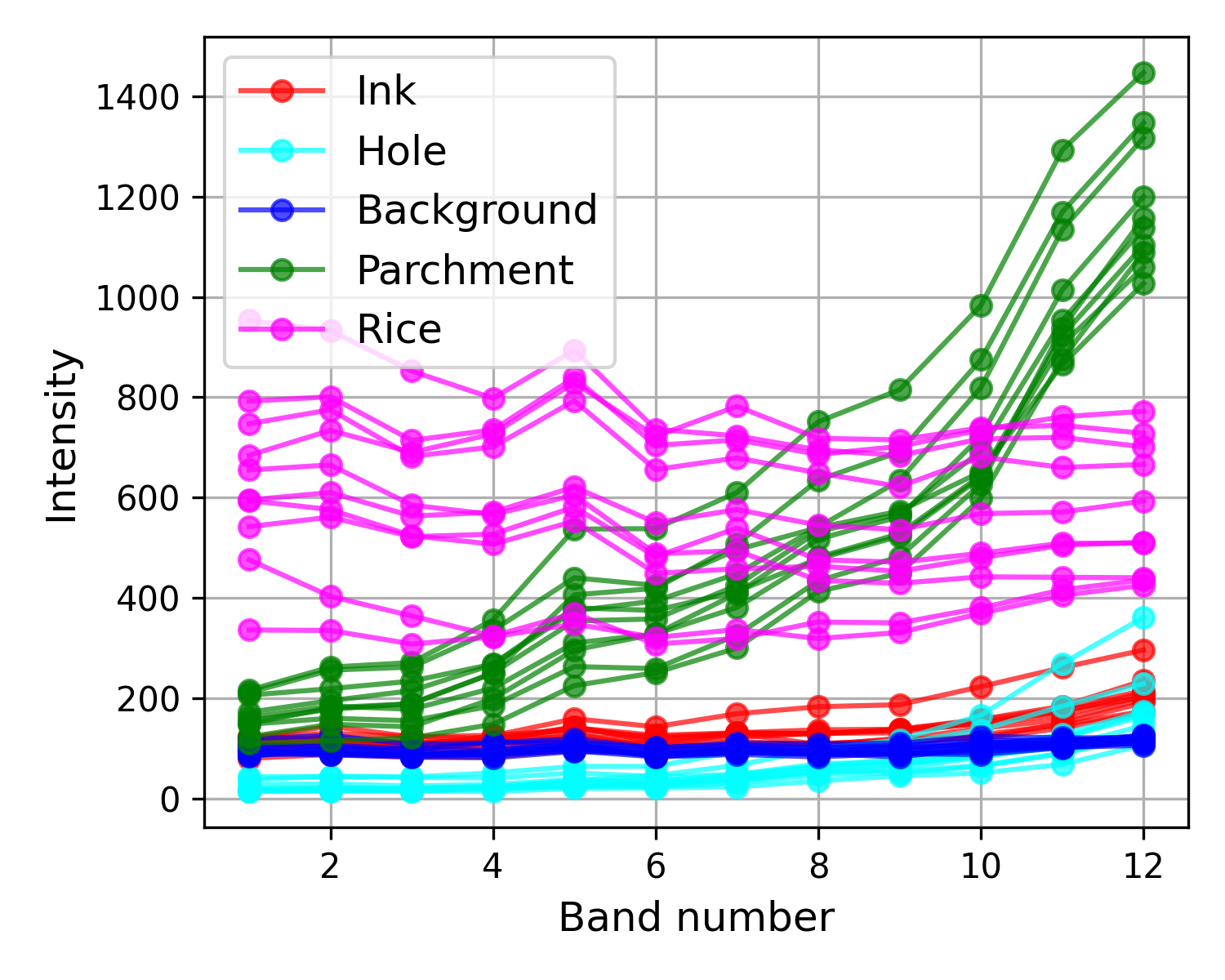}
  \caption{The spectral behavior of ink, parchment, background, hole, and rice materials within a DSS fragment across the 12 spectral bands. This analysis is useful for understanding the most discriminative bands for segmentation of ink and parchment regions}
  \label{fig:five_regions_ms_trends}
\end{figure}

\begin{figure*}[tb]
    \centering
    \begin{minipage}{0.38\textwidth}
        \centering
        \text{Images}
    \end{minipage}
    \begin{minipage}{0.38\textwidth}
        \centering
        \text{Histograms}
    \end{minipage}
    
    \begin{minipage}{0.03\textwidth}
        \centering
        \rotatebox{90}{Raw first-band}
    \end{minipage}
    \begin{minipage}{0.38\textwidth}
        \centering
        \includegraphics[width=\textwidth]{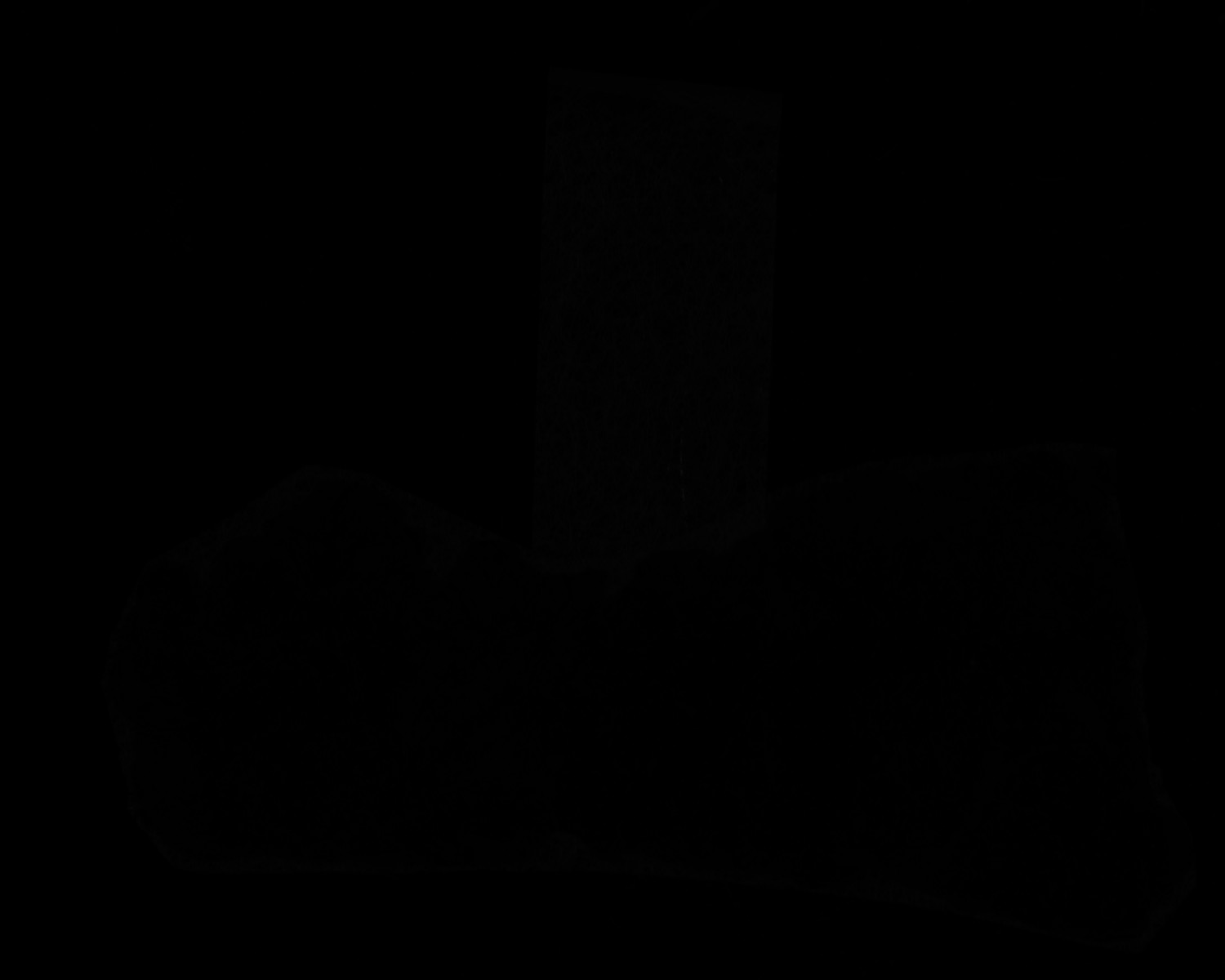}
    \end{minipage}
    \begin{minipage}{0.38\textwidth}
        \centering
        \includegraphics[width=\textwidth]{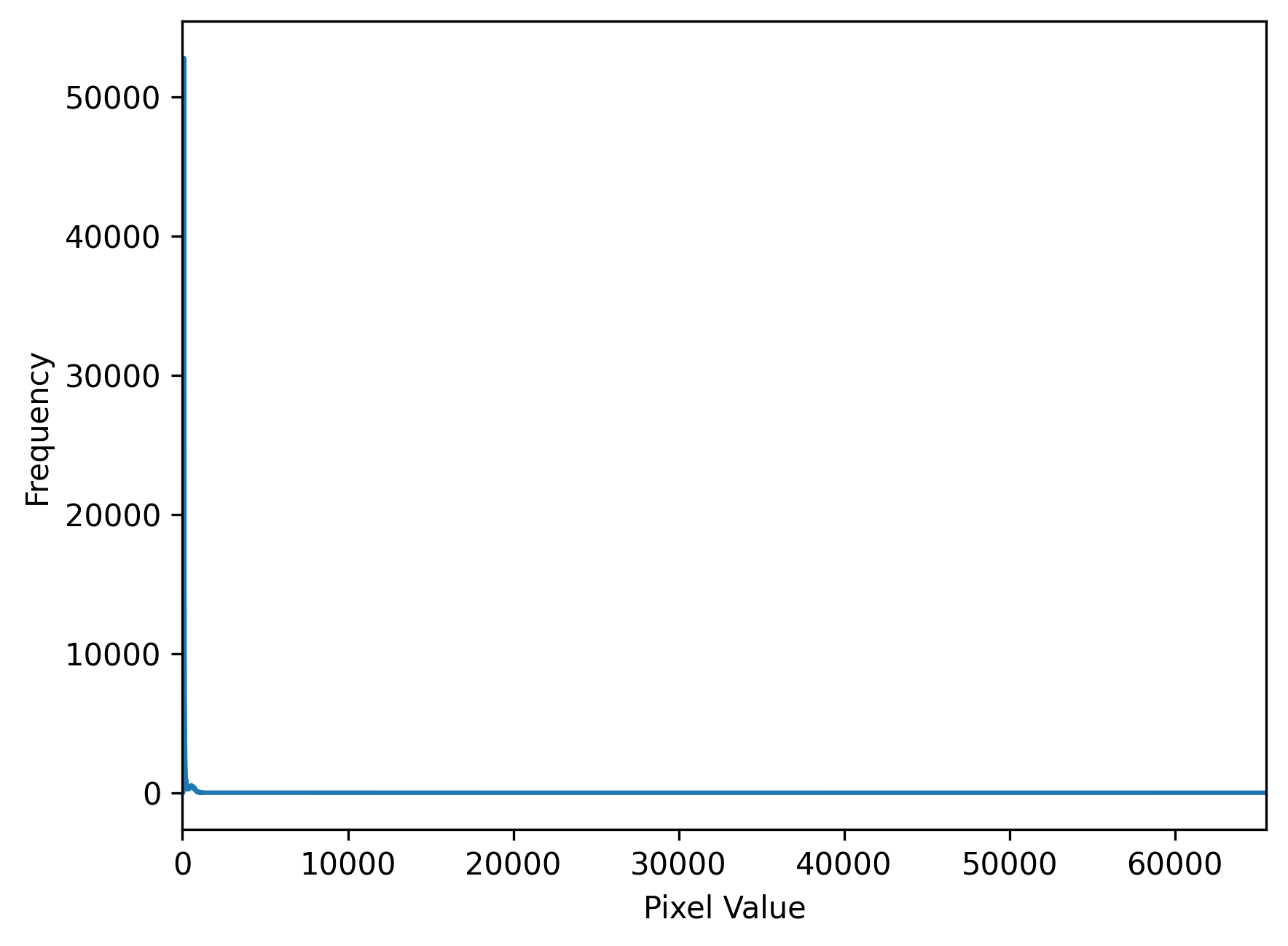}
    \end{minipage}

    \begin{minipage}{0.03\textwidth}
        \centering
        \rotatebox{90}{Normalized first-band}
    \end{minipage}
    \begin{minipage}{0.38\textwidth}
        \centering
        \includegraphics[width=\textwidth]{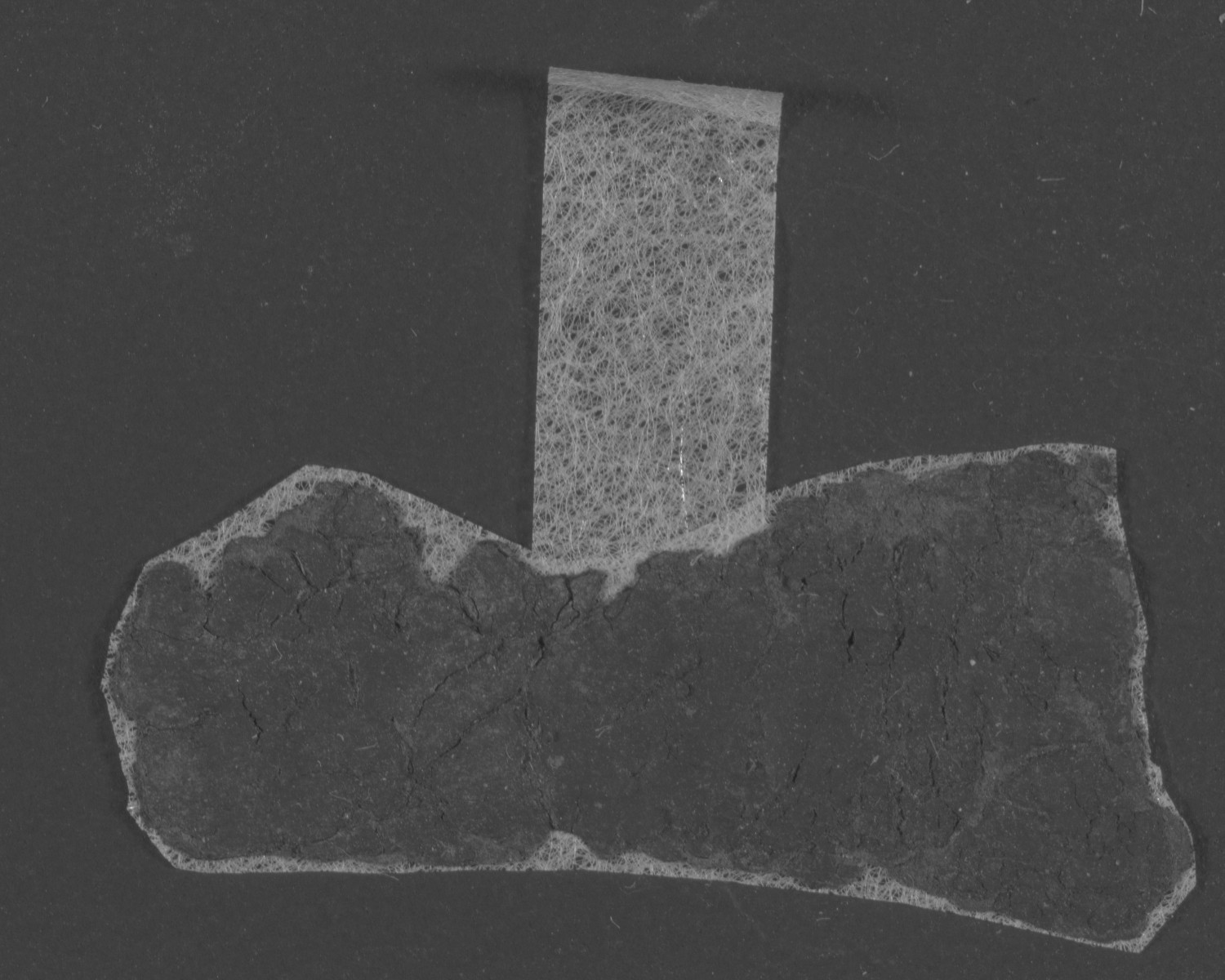}
    \end{minipage}
    \begin{minipage}{0.38\textwidth}
        \centering
        \includegraphics[width=\textwidth]{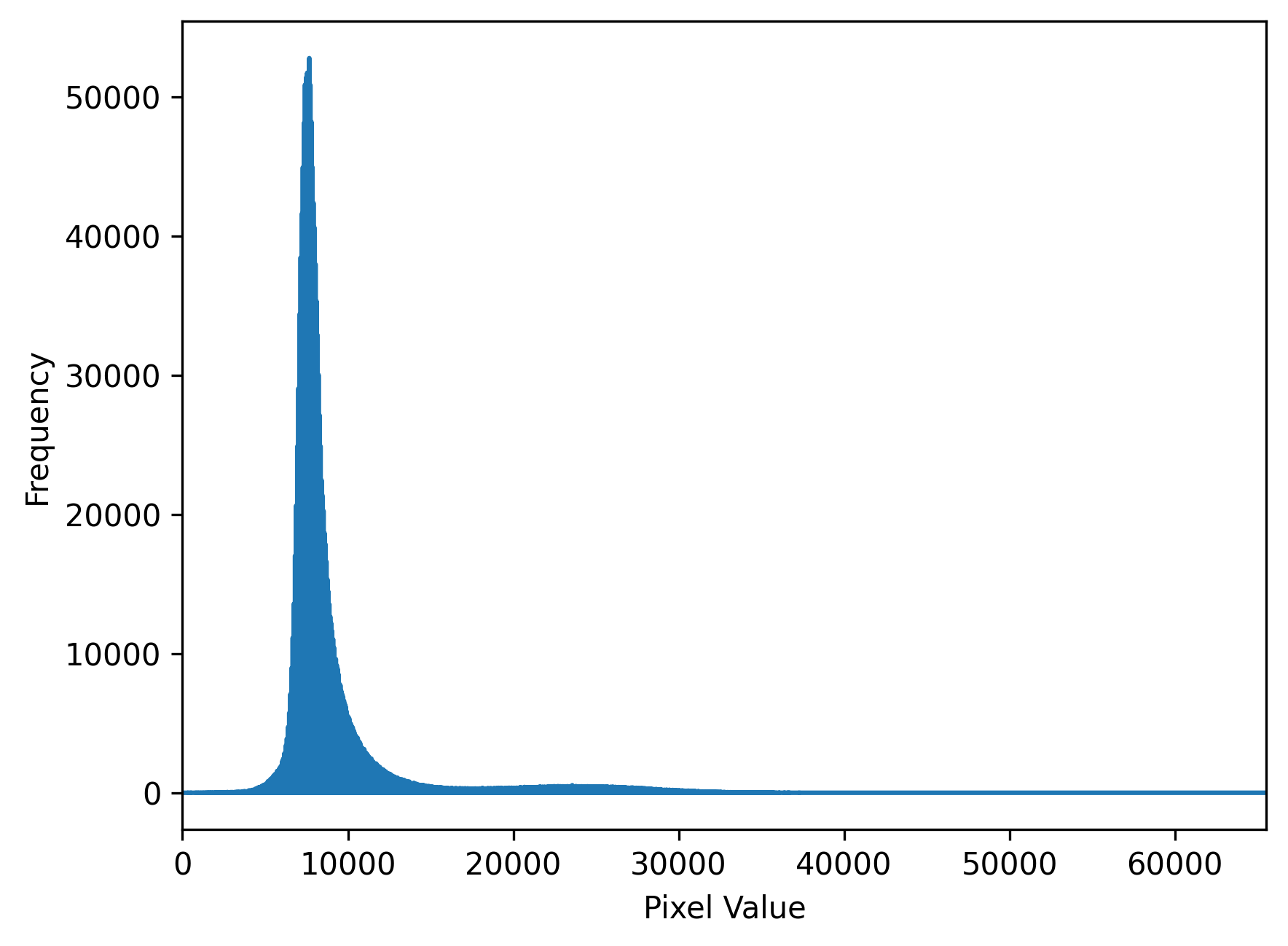}
    \end{minipage}

    \begin{minipage}{0.03\textwidth}
        \centering
        \rotatebox{90}{Raw last-band}
    \end{minipage}
    \begin{minipage}{0.38\textwidth}
        \centering
        \includegraphics[width=\textwidth]{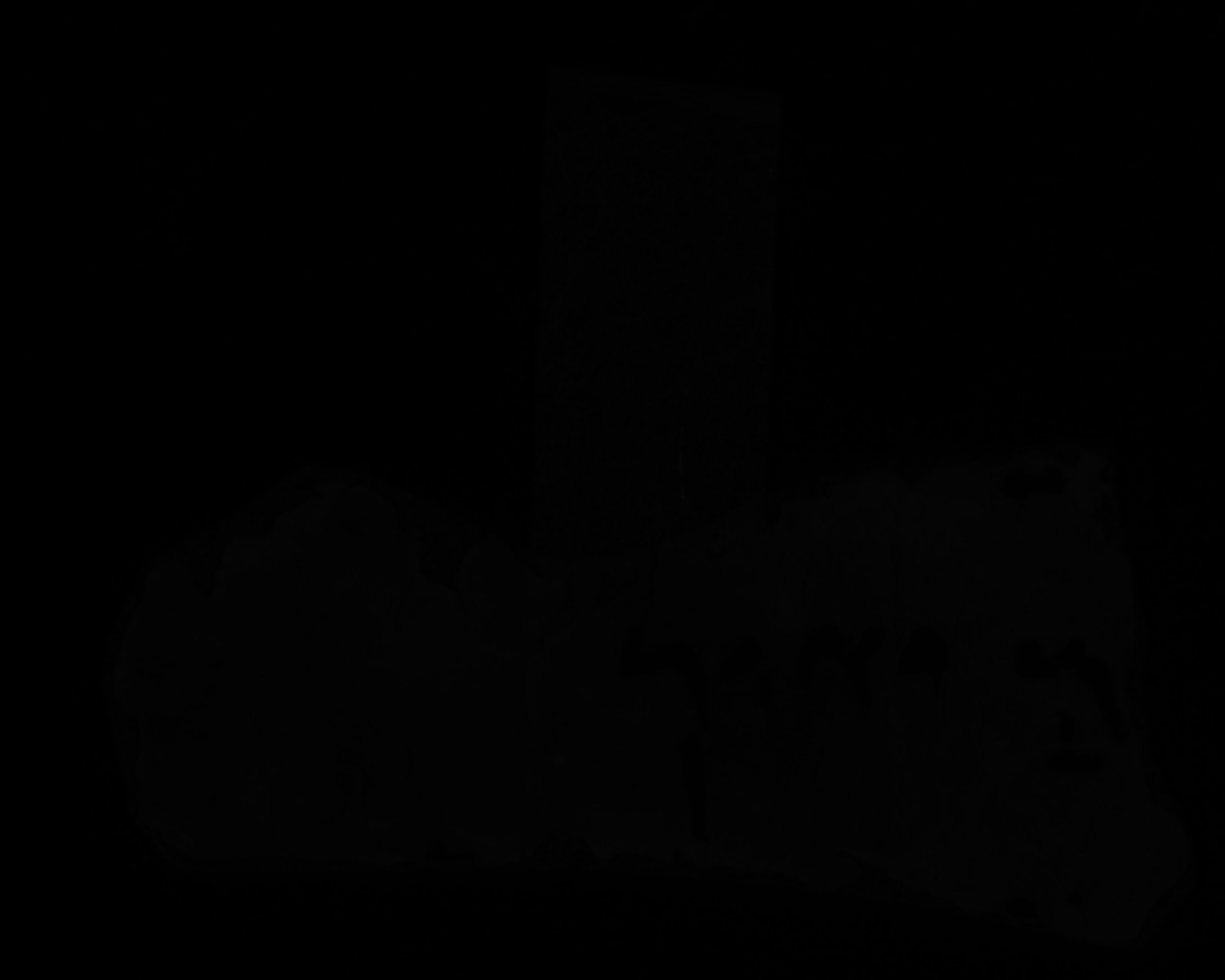}
    \end{minipage}
    \begin{minipage}{0.38\textwidth}
        \centering
        \includegraphics[width=\textwidth]{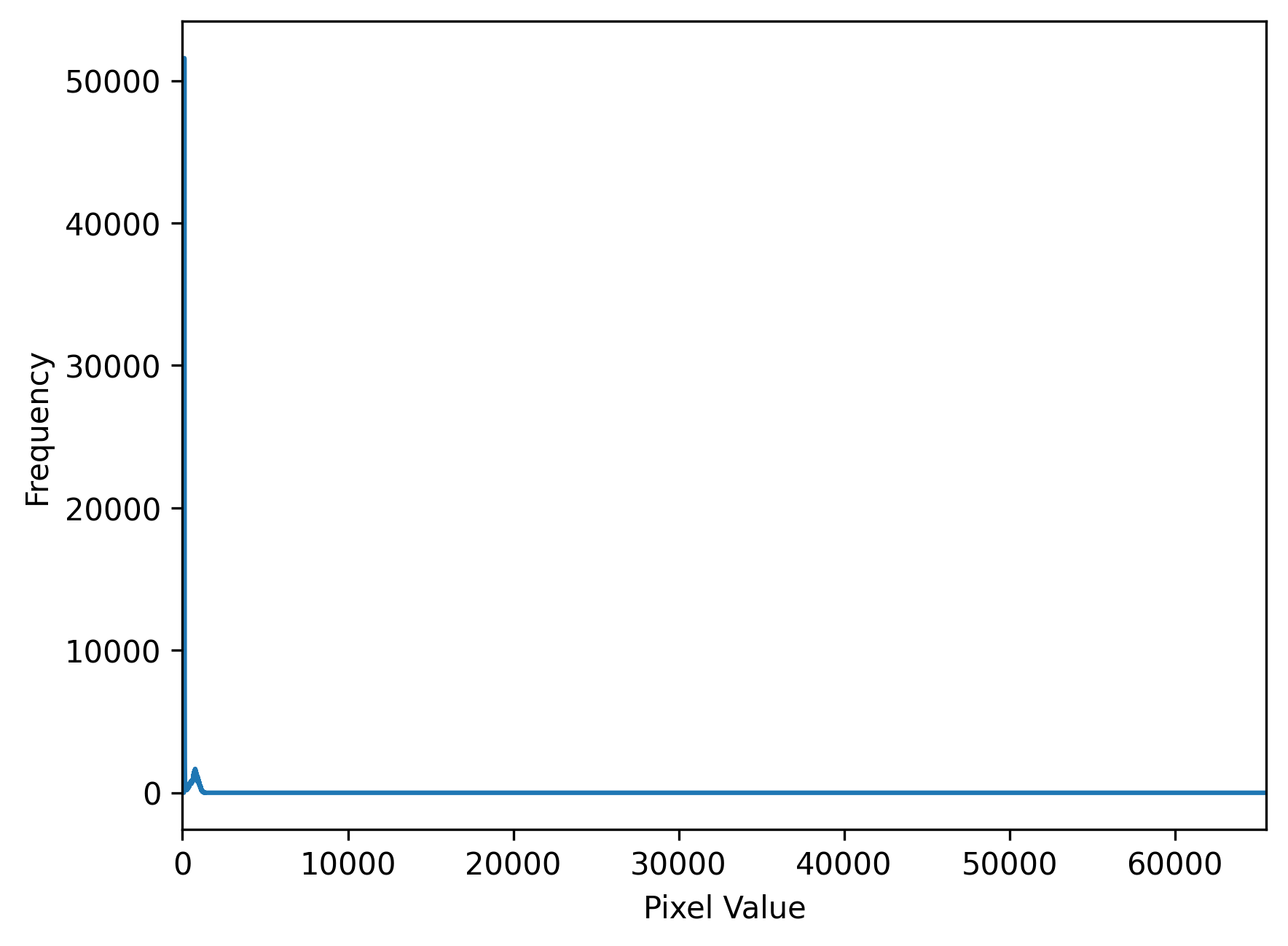}
    \end{minipage}

    \begin{minipage}{0.03\textwidth}
        \centering
        \rotatebox{90}{Normalized last-band}
    \end{minipage}
    \begin{minipage}{0.38\textwidth}
        \centering
        \includegraphics[width=\textwidth]{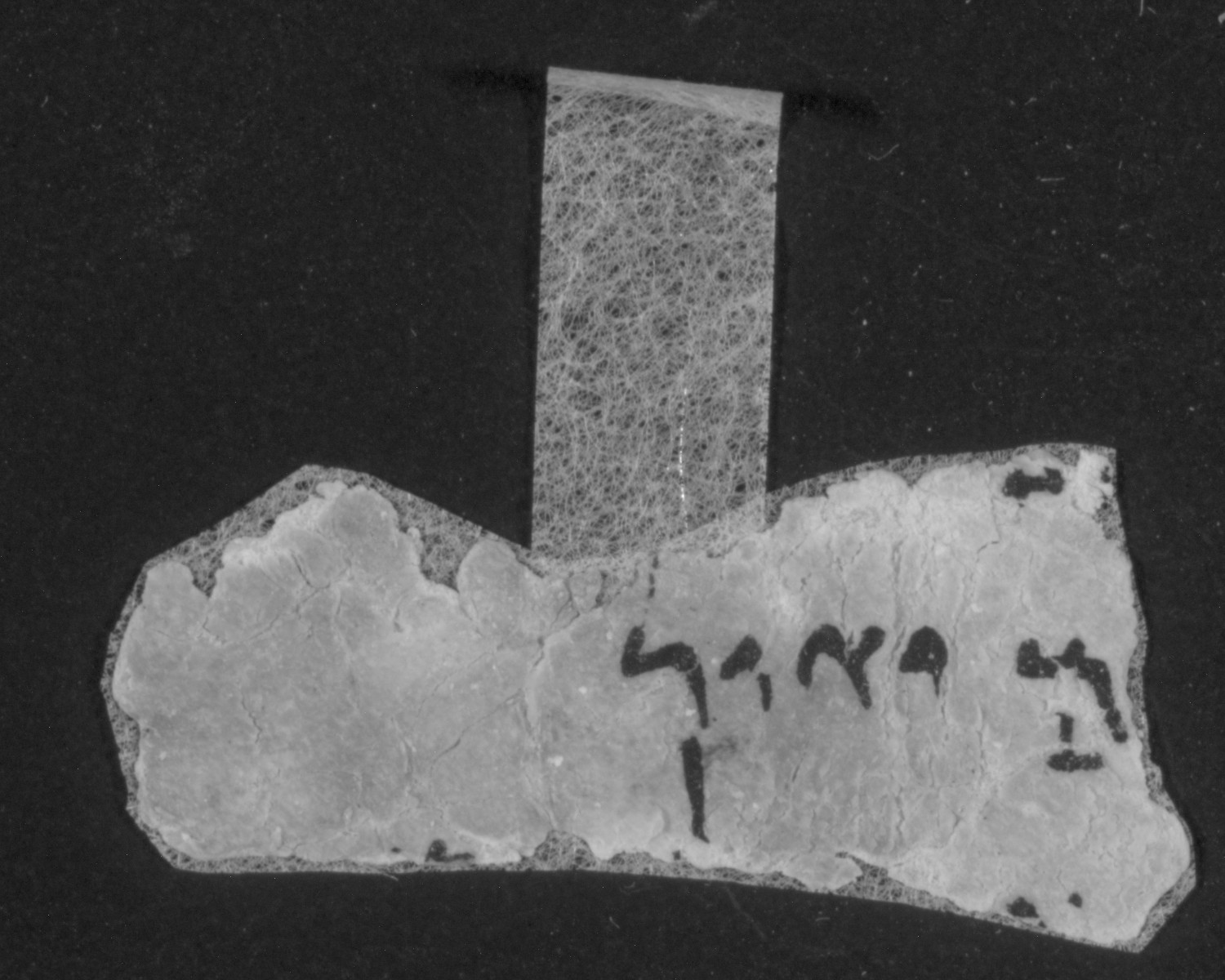}
    \end{minipage}
    \begin{minipage}{0.38\textwidth}
        \centering
        \includegraphics[width=\textwidth]{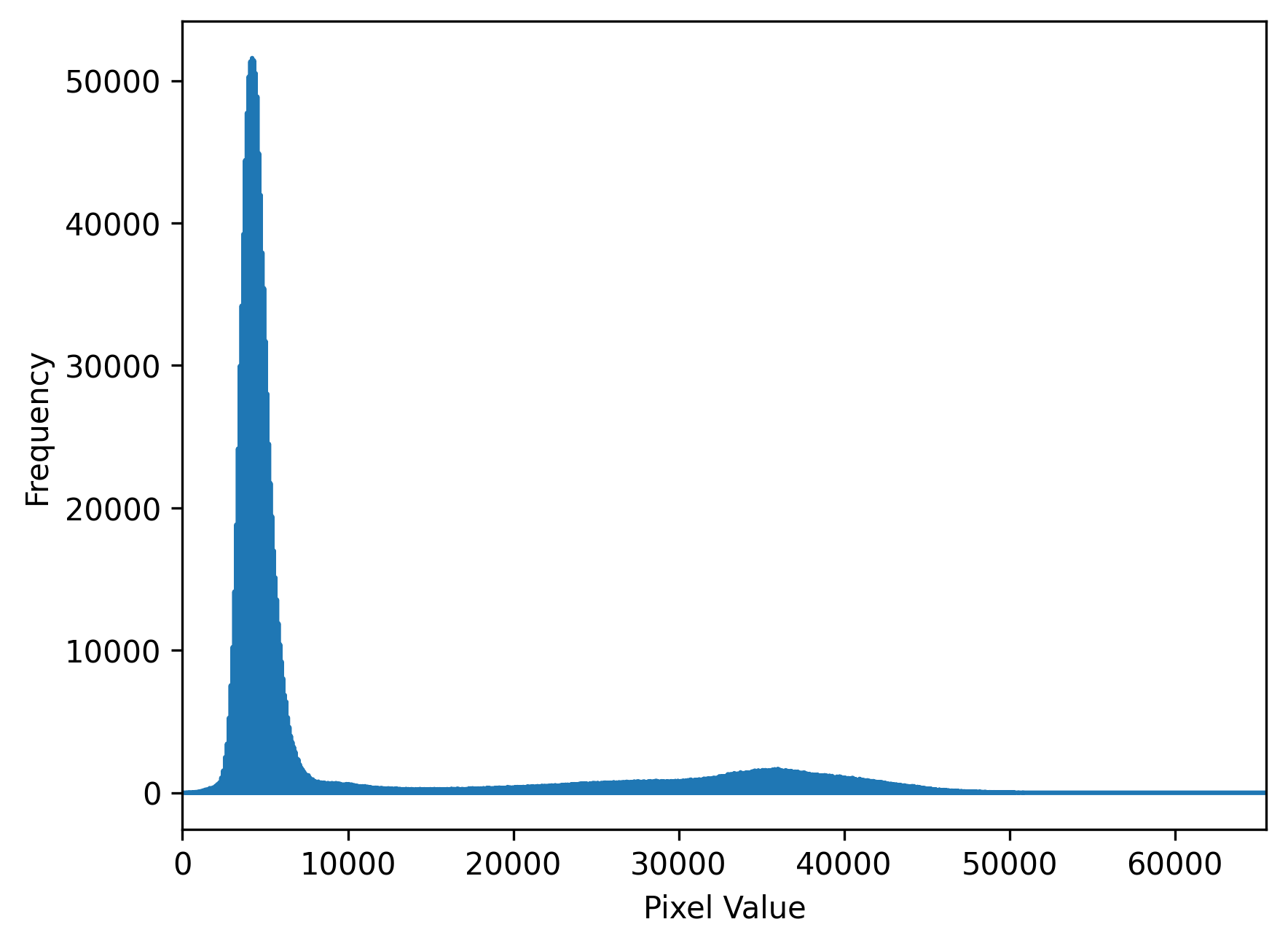}
    \end{minipage}
    \caption{The left column shows the raw 16-bit and normalized images for the first-band and last-band, while the right column displays their corresponding histograms. The raw images appear dark because the pixel values are concentrated in a narrow portion of the available range. Normalization expands these values, making features more visible and enhancing the overall contrast}
    \label{fig:histograms}
\end{figure*}

Typically, raw 16-bit first-band and last-band images appear dark because the pixel values cover a wide range, with most concentrated in a narrow portion. After normalization, we can see the ink is minimally visible in the first-band and maximally visible in the last-band (Figure~\ref{fig:histograms}). This is because the parchment has low reflectivity in the blue band and high reflectivity in the near-infrared bands, while the ink remains consistently low in reflectivity across all bands, with a slight increase in the near-infrared bands. This slight increase occurs because near-infrared light penetrates beneath the ink layer and reflects off the underlying parchment (Figure~\ref{fig:five_regions_ms_trends}).

\subsubsection{Normalization}

The full-color images and normalized last-band images were included in the dataset for qualitative evaluation and potential future work. The full-color image reveals details without requiring additional processing. In contrast, the raw 16-bit last-band image, despite having the highest contrast between the ink and parchment, does not appear clearly to the eye because its pixel values are concentrated in a narrow range relative to the full span of 0--65535. Therefore, a normalized version of the last-band images was included to enhance visibility. 

Since the raw images are stored as integer intensity values, gamma encoding was applied before normalization. Let $I(x,y)$ denote the pixel intensity at location $(x,y)$ in the raw 16-bit last-band image. The image is first scaled to the unit range $[0, 1]$:
\[
I_{\text{unit}}(x,y) = \frac{I(x,y)}{65535}
\]
Next, gamma encoding is applied using the equation:
\[
I_{\text{gamma}}(x,y) = I_{\text{unit}}(x,y)^{1/\gamma}
\]
 where $\gamma = 2.2$, which is a standard choice. After gamma encoding, the resulting image is normalized to the range $[0, 65535]$:
\[
I_{\text{normal}}(x,y) = 65535 \cdot \frac{I_{\text{gamma}}(x,y) - \min(I_{\text{gamma}})}{\max(I_{\text{gamma}}) - \min(I_{\text{gamma}})}
\]
Finally, the image is converted back to a 8-bit format and saved as a JPEG file for ease of use.

\section{Method}
In this section, we present the MTEM method for segmenting ink and parchment regions in high-resolution images of DSS fragments. Recognizing the challenges inherent in manual annotation due to the intricate details and physical deformations of the fragments, our approach leverages multispectral thresholding and energy minimization techniques to differentiate between ink, parchment, and other regions such as holes, background, and rice paper. By analyzing the spectral characteristics across multiple bands, we identify distinctive intensity trends that enable us to apply thresholding methods effectively. Subsequently, we refine the segmentation through energy minimization, resulting in accurate delineation of ink and parchment.

\subsection{Method motivation}

\begin{figure*}[htb]
  \centering
  \includegraphics[width=.95\linewidth]{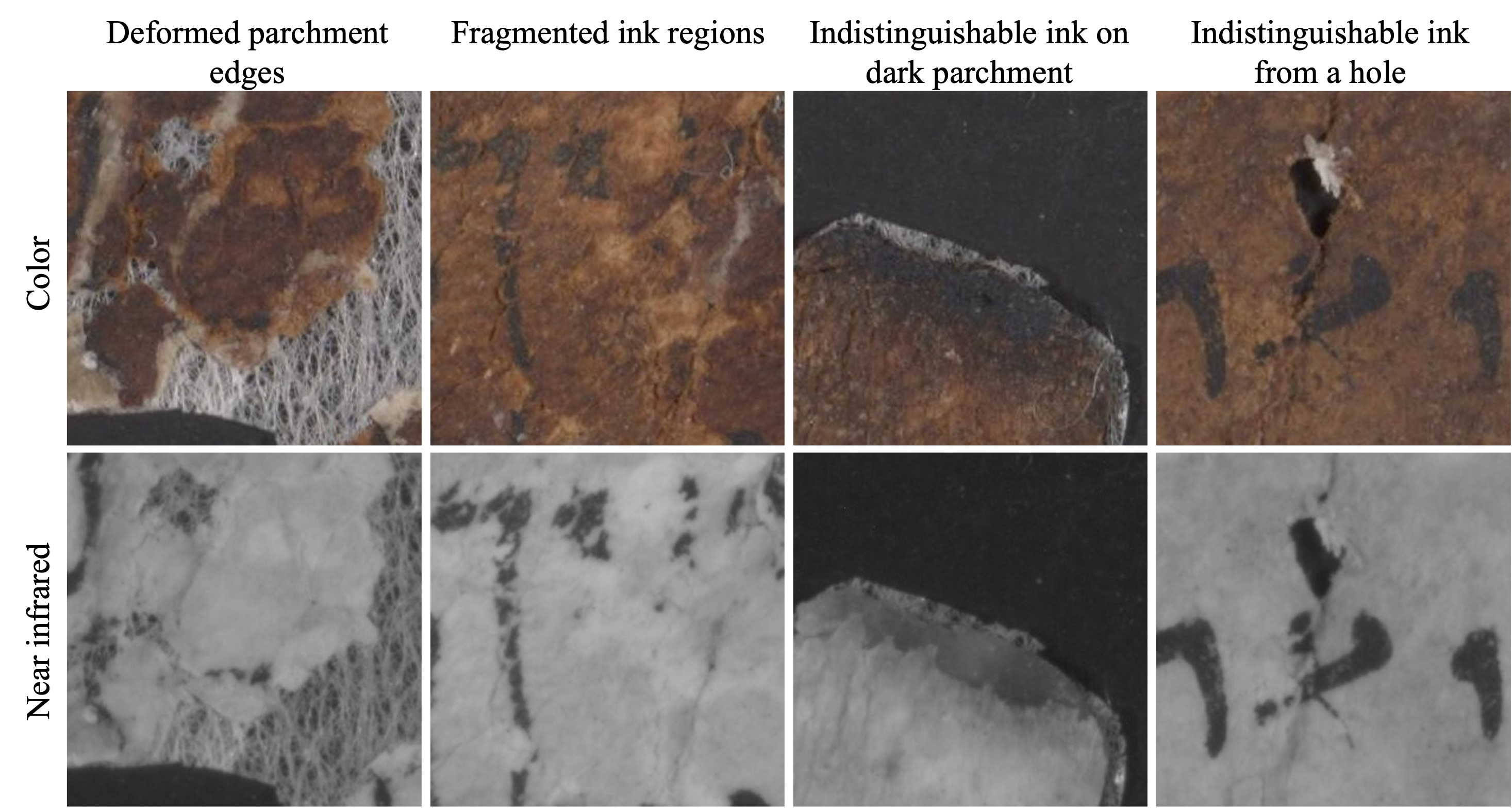}
  \caption{DSS fragment image patches illustrating various annotation challenges. The first row shows color images highlighting deformed parchment edges, fragmented ink regions, indistinguishable ink on dark parchment, and ink that is difficult to differentiate from a hole. The second row displays corresponding near-infrared images, which, while enhancing ink visibility on normal parchment, fail to reveal clear ink edges on darkened parchment and do not help to differentiate ink from holes.}
  \label{fig:challenges}
\end{figure*}

Annotating DSS fragments is a challenging task due to the high resolution of the images ($7216\times 5412$ pixels), which capture intricate and important details that may require expert interpretation. Additionally, the physical nature of the fragments adds complexity to the annotation process. The edges of the parchment are often highly irregular and deformed, making precise delineation difficult. Ink regions are frequently fragmented, or faded, which significantly increases the length and complexity of the edges to be segmented. Distinguishing between ink and parchment becomes even more challenging on darkened areas of the parchment. Moreover, holes in the parchment are particularly difficult to differentiate from ink regions due to their similar visual characteristics. While near-infrared images make ink more visible on normal parchment, they are not sufficiently helpful for discriminating between ink and holes or revealing the clear edges of ink on darkened parchment (Figure~\ref{fig:challenges}).

Given these challenges, we aimed to develop a method that bypasses the need for manual and expert-driven annotation, providing an automated solution to aid in the analysis of DSS fragments.

\subsection{Multispectral characterization} 

Given a fragment image, our goal is to segment the ink and parchment regions. In other words, we aim to identify the pixels corresponding to the ink and those corresponding to the parchment. However, a fragment image also contains additional regions beyond ink and parchment, such as rice, background, and holes. While we do not need to differentiate between the rice, background, and hole regions, it is crucial to distinguish them from the ink and parchment regions. To identify the most discriminative bands or band differences for segmentation of ink and parchment, we randomly sample pixels from each of the five region types—ink, parchment, rice, background, and holes—and analyze their multispectral intensity trends.

Figure~\ref{fig:five_regions_ms_trends} illustrates the multispectral intensity trends for each region across the 12 spectral bands. We observe that parchment exhibits a highly distinctive trend from holes, background and rice, marked by significantly higher reflectivity in the near-infrared bands. Rice also shows a unique trend that remains relatively consistent, with minimal overlap with the intensity values of ink. In contrast, the trends for ink, holes, and background are more similar and tend to overlap, displaying small variances. Background pixels maintain a consistent intensity across all bands but have similar values to those of ink. Hole intensities are closely grouped in the initial bands but increase in the final bands, where they start to overlap with ink, which also shows an increase in intensity. The reflectivity increase in ink regions is attributed to near-infrared light penetrating the ink layer and scattering off the underlying parchment. The reflectivity increase in hole regions is due to diffusive reflection of near-infrared light from the surrounding parchment edges near the hole. 

\begin{figure*}[htb]
    \centering
    \includegraphics[width=0.3\textwidth]{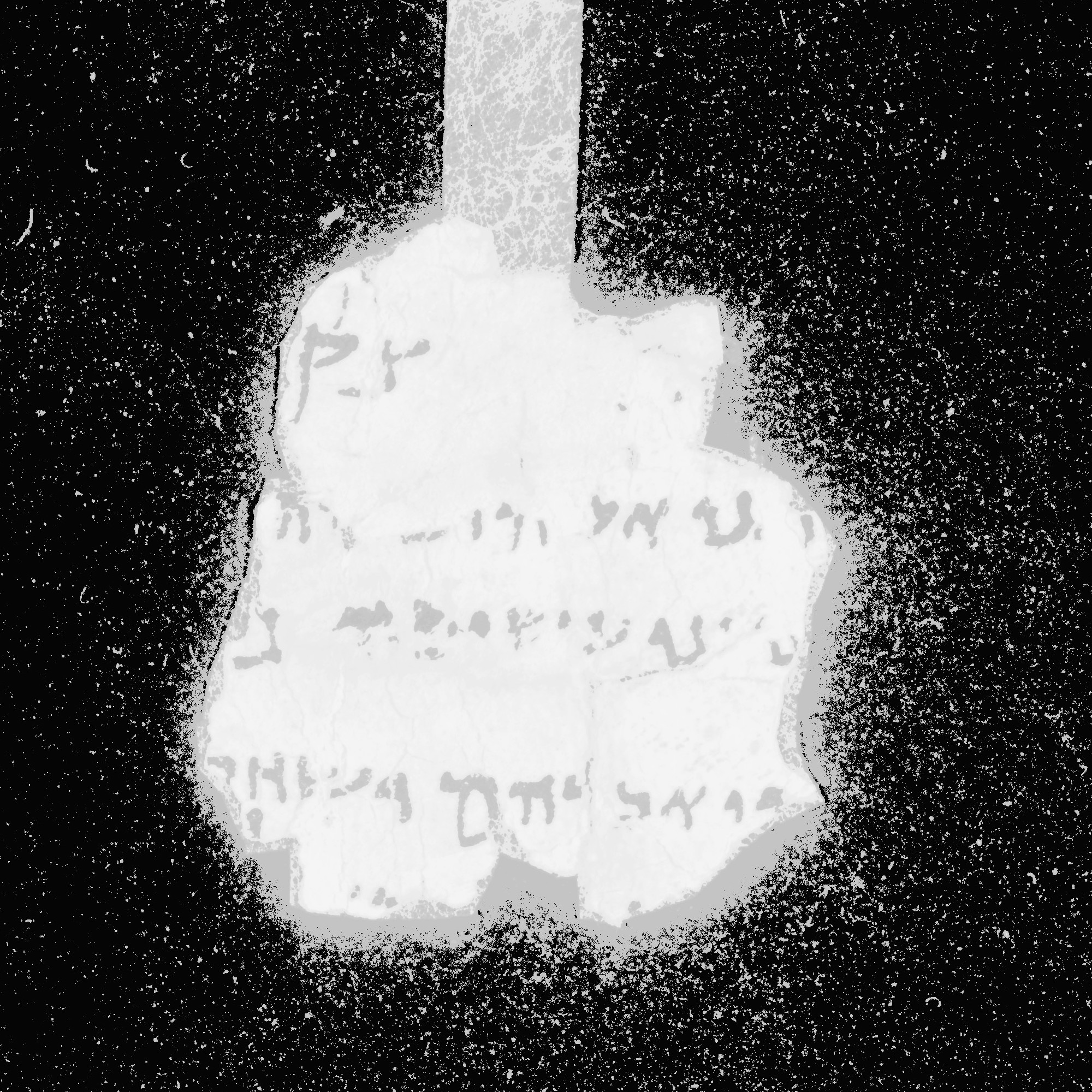}
    \includegraphics[width=0.3\textwidth]{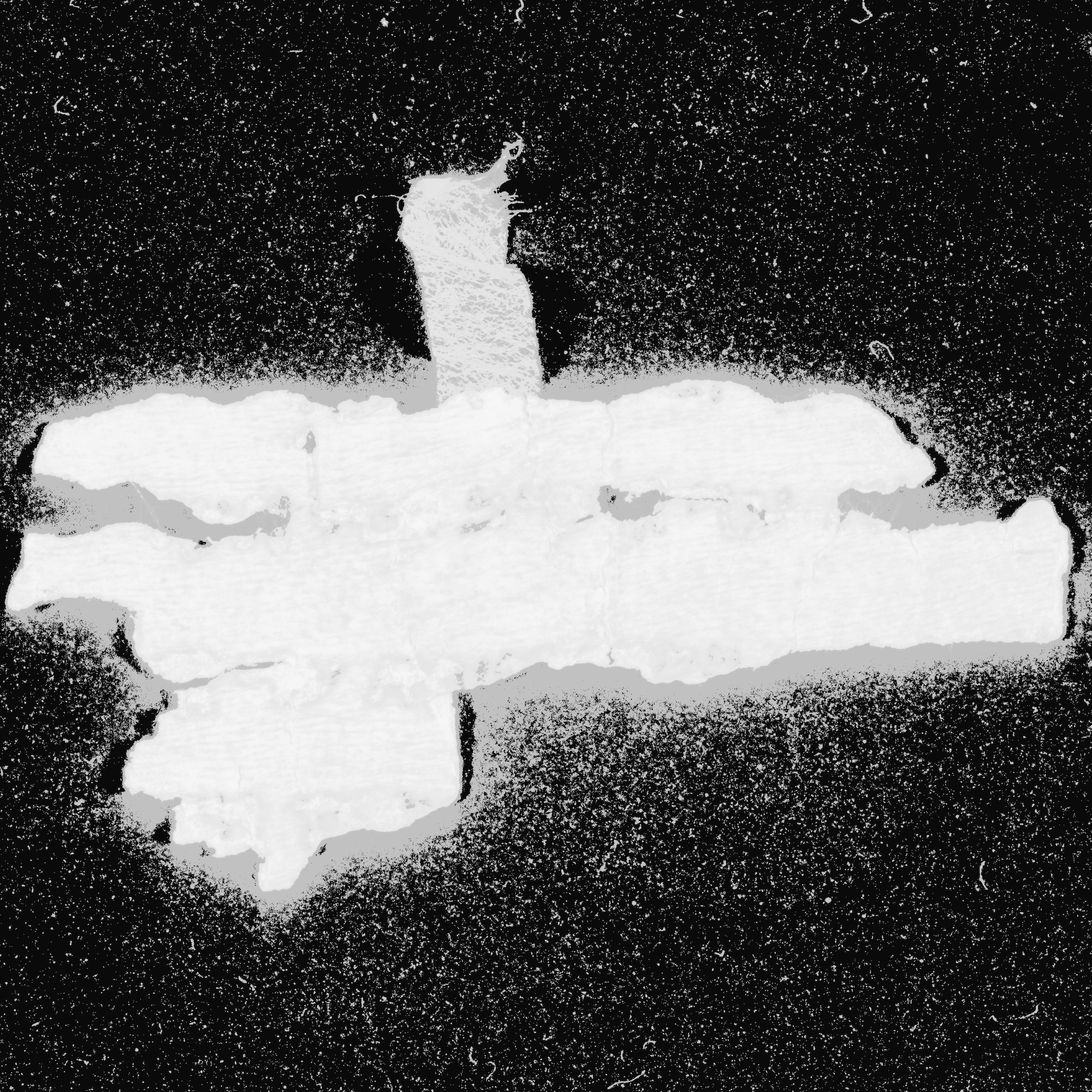}
    \includegraphics[width=0.3\textwidth]{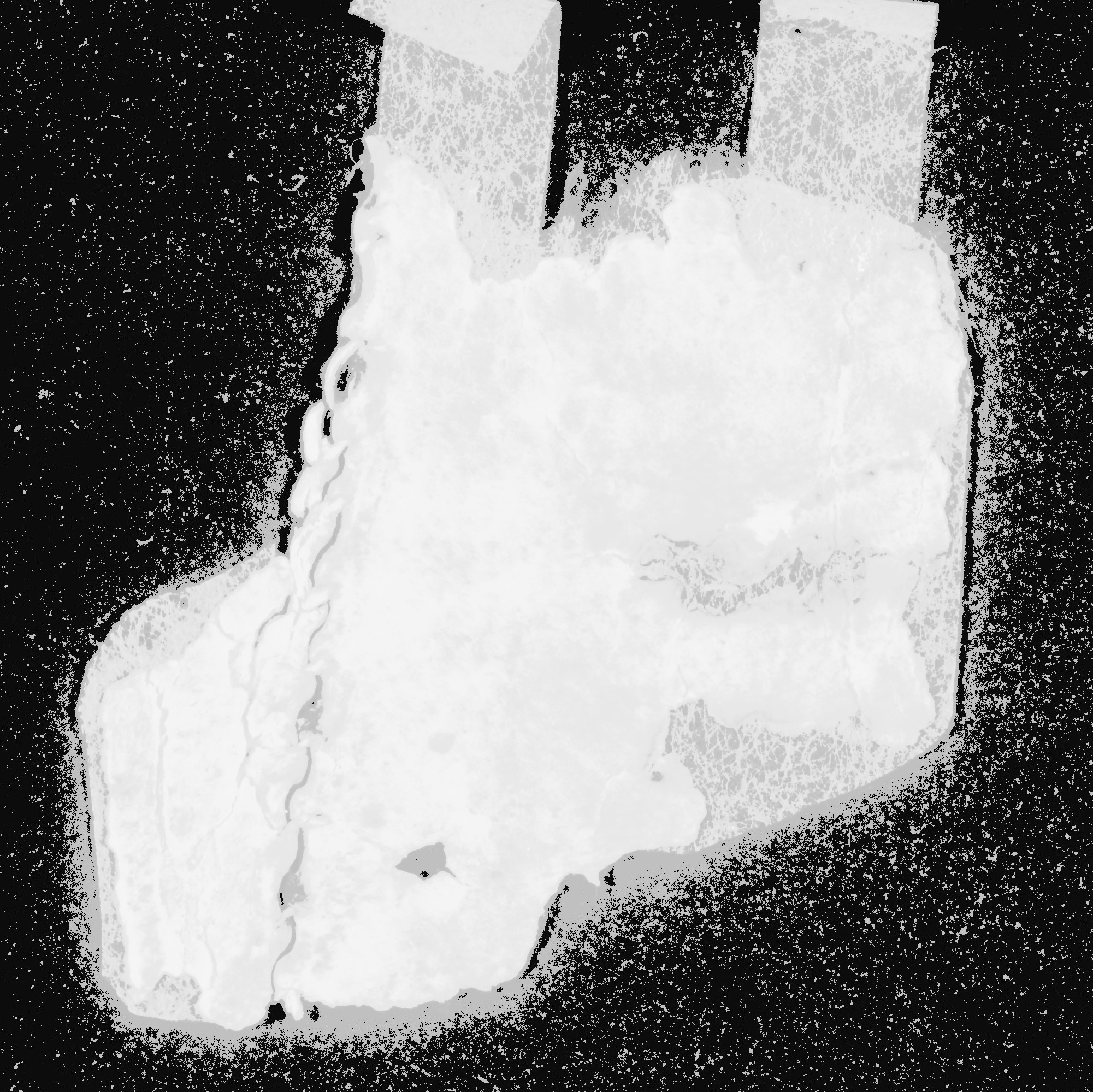}
    \caption{Equalized last-band images of three DSS fragments showing halos at the edges of the parchment. We hypothesize that the halos and holes are background regions influenced by diffusive reflections from the parchment in near-infrared bands, where the higher reflectivity of parchment makes these effects prominent}
    \label{fig:halo}
\end{figure*}

We hypothesize that the diffusive reflectivity also contributes to halos in the IAA images, which become particularly prominent in the near-infrared bands (Figure~\ref{fig:halo}). This is due to the higher reflectivity of parchment in those bands, making the halos visible in background areas adjacent to parchment regions. This observation also supports the reasoning regarding the holes, explaining why they appear to have slightly higher reflectivity than the background: they are surrounded by parchment and affected by strong diffusive reflections. Thus, for holes to be classified as such, they should remain relatively small; beyond a certain size, they start behaving more like the background.

\begin{figure}[htb]
  \centering
  \includegraphics[width=.95\linewidth]{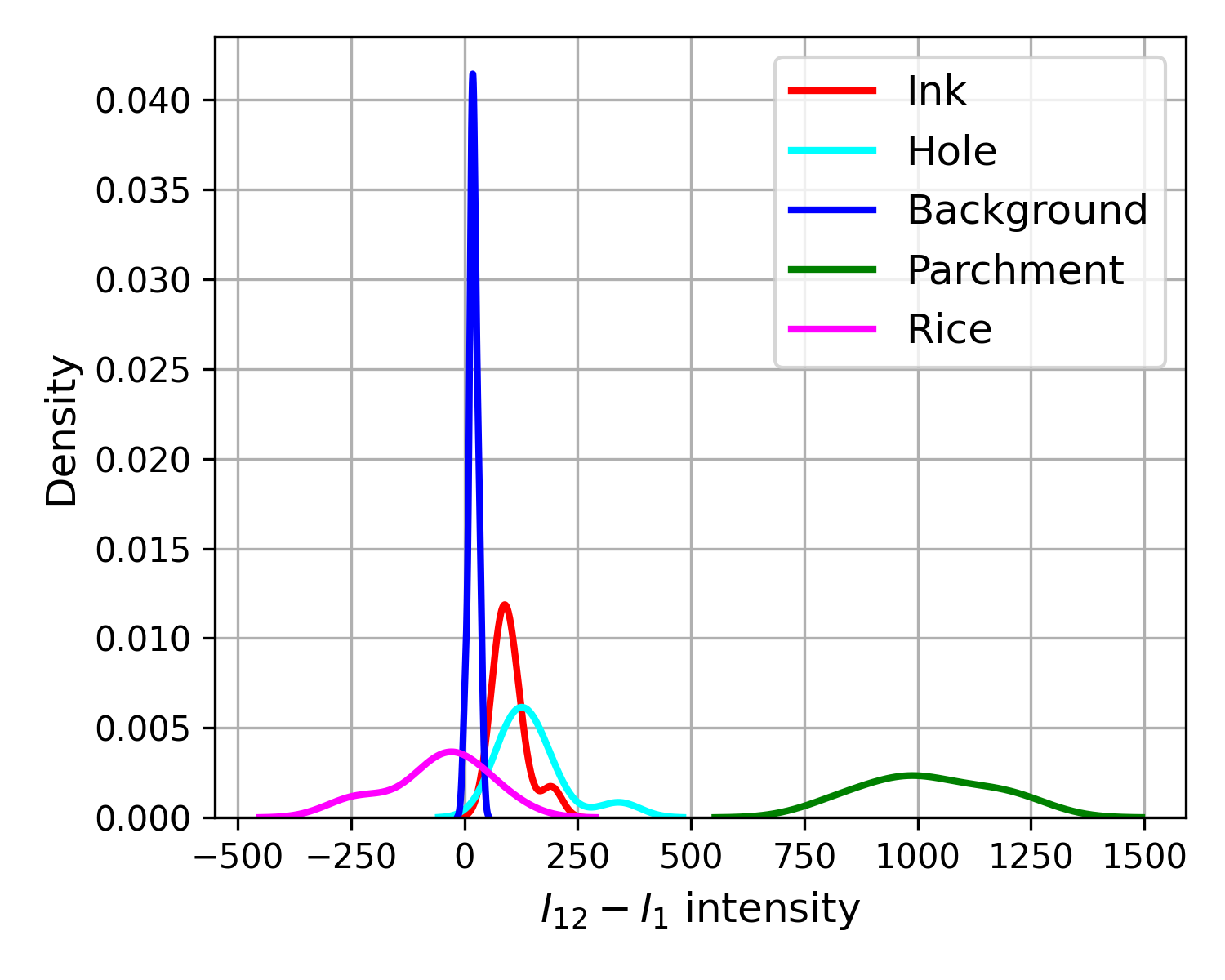}
  \caption{Intensity distribution of \(I_{12} - I_1\) across different regions, illustrating the separability of parchment pixels from other regions}
  \label{fig:band12_minus_band1_intensity_distribution}
\end{figure}

\begin{figure}[htb]
  \centering
  \includegraphics[width=.95\linewidth]{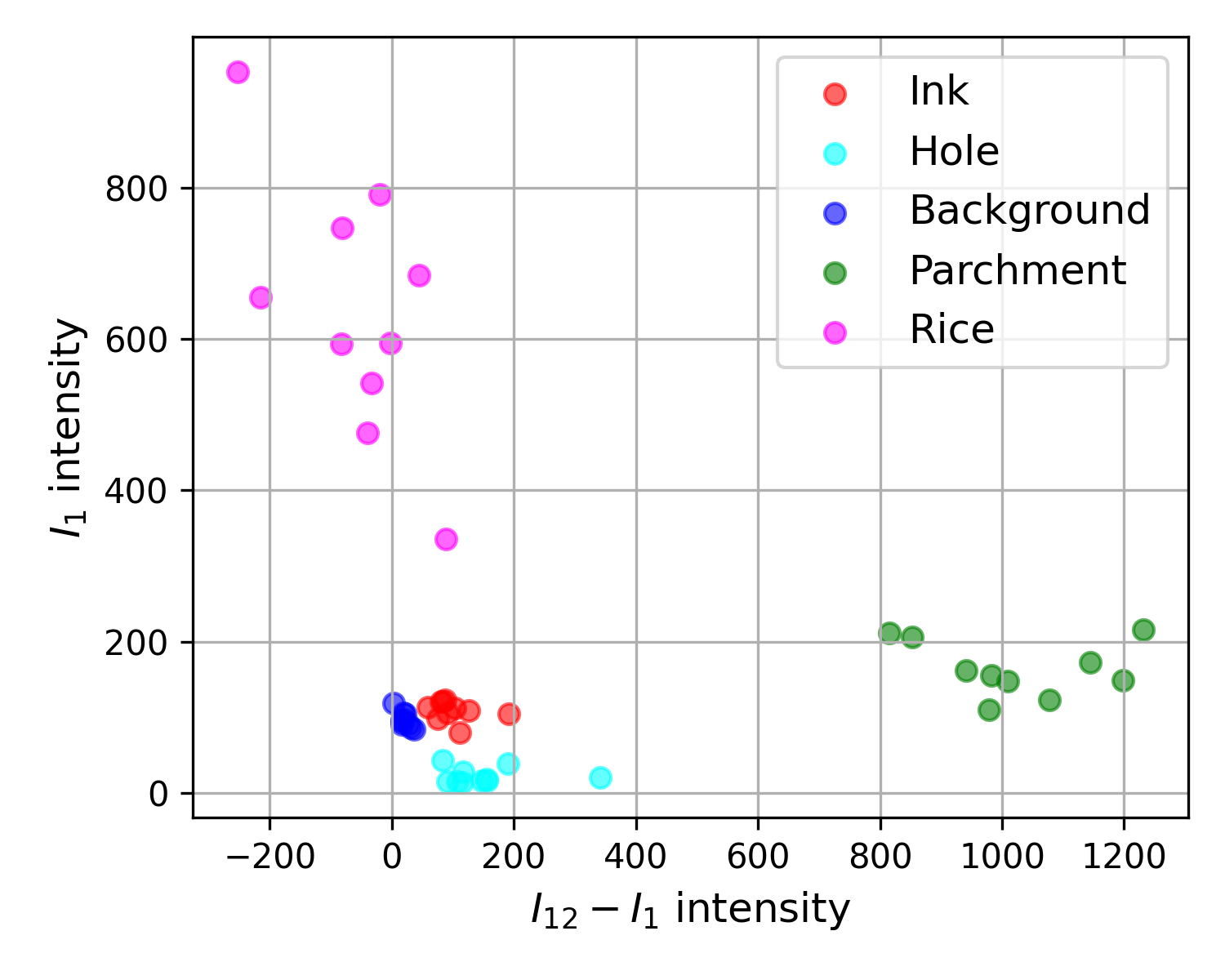}
  \caption{Combination of \(I_1\) and \(I_{12} - I_1\) values across different regions, illustrating the separability of ink pixels from other regions}
  \label{fig:band12_minus_band1_versus_band_1_intensity_distribution}
\end{figure}

Figure~\ref{fig:five_regions_ms_trends} reveals that parchment pixels achieve the best discrimination using the intensity difference between band 12 and band 1, denoted as \(I_{12} - I_1\), where \(I_n\) represents the image of band \(n\). This band difference emphasizes the uniquely increasing trend of parchment in contrast to all other regions, as shown in Figure~\ref{fig:band12_minus_band1_intensity_distribution}. Similarly, ink pixels are best distinguished when using a combination of \(I_{12} - I_1\) and \(I_1\), where \(I_1\) helps to exclude hole and rice pixels, while the \(I_{12} - I_1\) difference helps to exclude parchment and background pixels (Figure~\ref{fig:band12_minus_band1_versus_band_1_intensity_distribution}).

\subsection{Multispectral thresholding} \label{sec:multispectral_thresholding}

After identifying that parchment can be distinguished using the intensity difference \(I_{12} - I_1\) and ink can be distinguished using a combination of \(I_{12} - I_1\) and \(I_1\), we employed annotations from the fragment 124-001 to set thresholds for differentiating parchment and ink from other regions. For each of the ink and parchment regions, we applied a percentile-based method using region-specific intensity values from the annotated fragment. Percentiles allow us to set thresholds by focusing on the central distribution of pixel intensities, reducing the influence of extreme values or noise. The \(n\)-th percentile, denoted as \(P_n\), is the value below which \(n\%\) of the data falls. Accordingly, we defined the lower threshold for each region as its \(P_n\) value and the upper threshold as its \(P_{100 - n}\) value, thus capturing the primary intensity range while excluding outliers.

For thresholding parchment, we applied the \(n\)-th percentile on \(I_{12} - I_1\)  because multispectral characterization indicated that \(I_{12} - I_1\)  provides a discrimination between parchment and other regions.

For thresholding ink, we applied the \(n\)-th percentile on both \(I_1\) and \(I_{12} - I_1\), as multispectral characterization demonstrated that the combination of \(I_1\) and \(I_{12} - I_1\) can distinguish ink from other regions.

\begin{figure*}[htb]
  \centering
  \includegraphics[width=.95\linewidth]{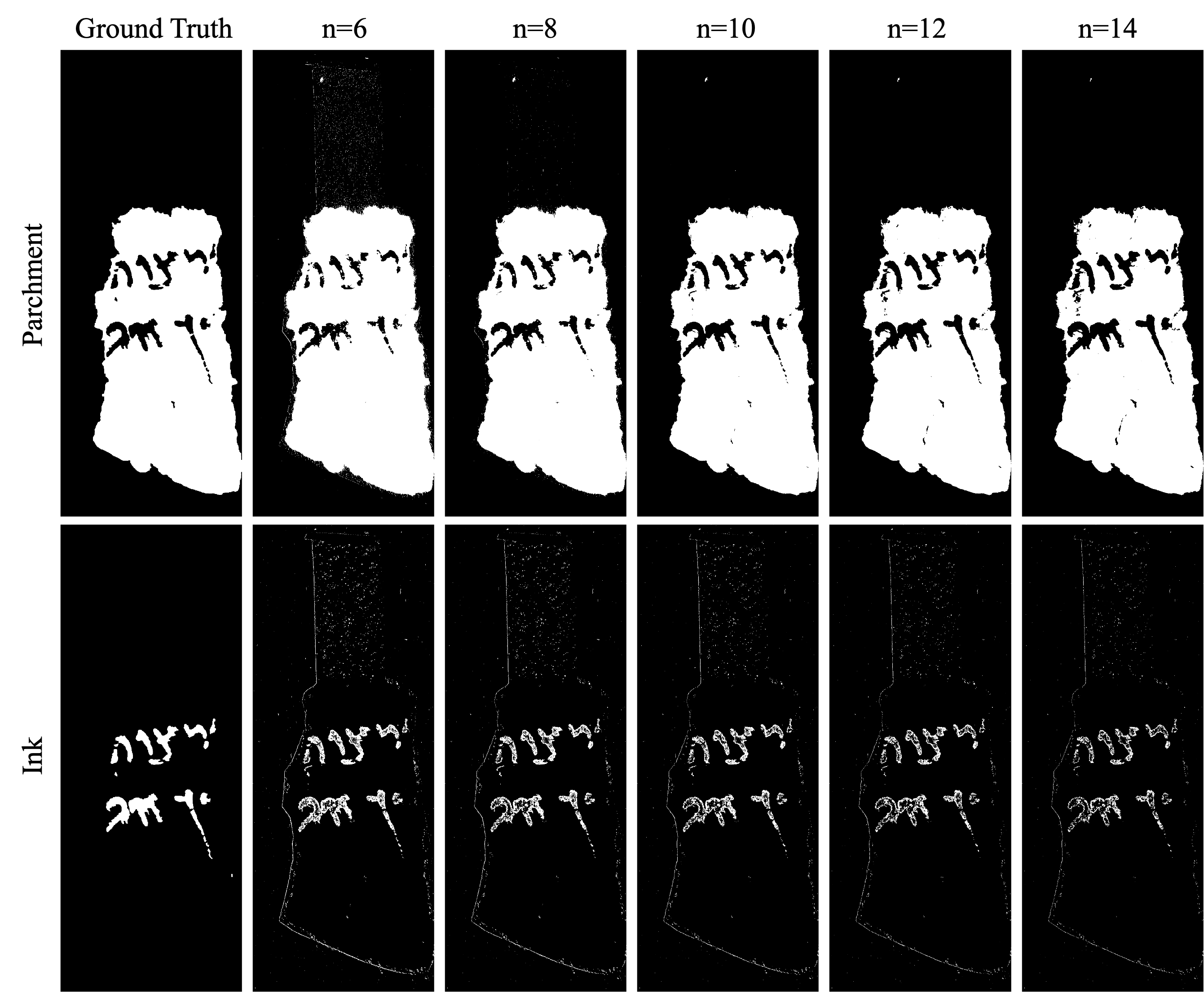}
  \caption{Multispectral thresholding results of parchment and ink regions with varying percentiles with the first column as ground truth}
  \label{fig:n_versus_parchments_and_inks}
\end{figure*}

Figure~\ref{fig:n_versus_parchments_and_inks} illustrates that as the percentile threshold decreases, segmentation for both ink and parchment becomes more complete but also incorporates more noise from other regions; conversely, as the percentile increases, segmentation becomes less complete but contains less noise. Based on experimental assessment in Section~\ref{sec:optimal_percentile}, we chose \(n=10\) for parchment thresholding that includes the maximum parchment pixels with the least noise pixels. However, the segmented parchment alone is not the primary focus; scholars are more interested in the combined parchment $\cup$ ink regions, which requires accurate ink segmentation for effective parchment segmentation.

Notably, none of the tested \(n\) values yield satisfactory ink segmentation. Ink regions encompass not only ink but also cracks, fading, holes, and degraded parchment. Based on experimental assessment in Section~\ref{sec:optimal_percentile}, we chose \(n=10\) for ink thresholding that includes the maximum ink pixels with the least noise pixels.

\begin{figure}[htb]
  \centering
  \includegraphics[width=.95\linewidth]{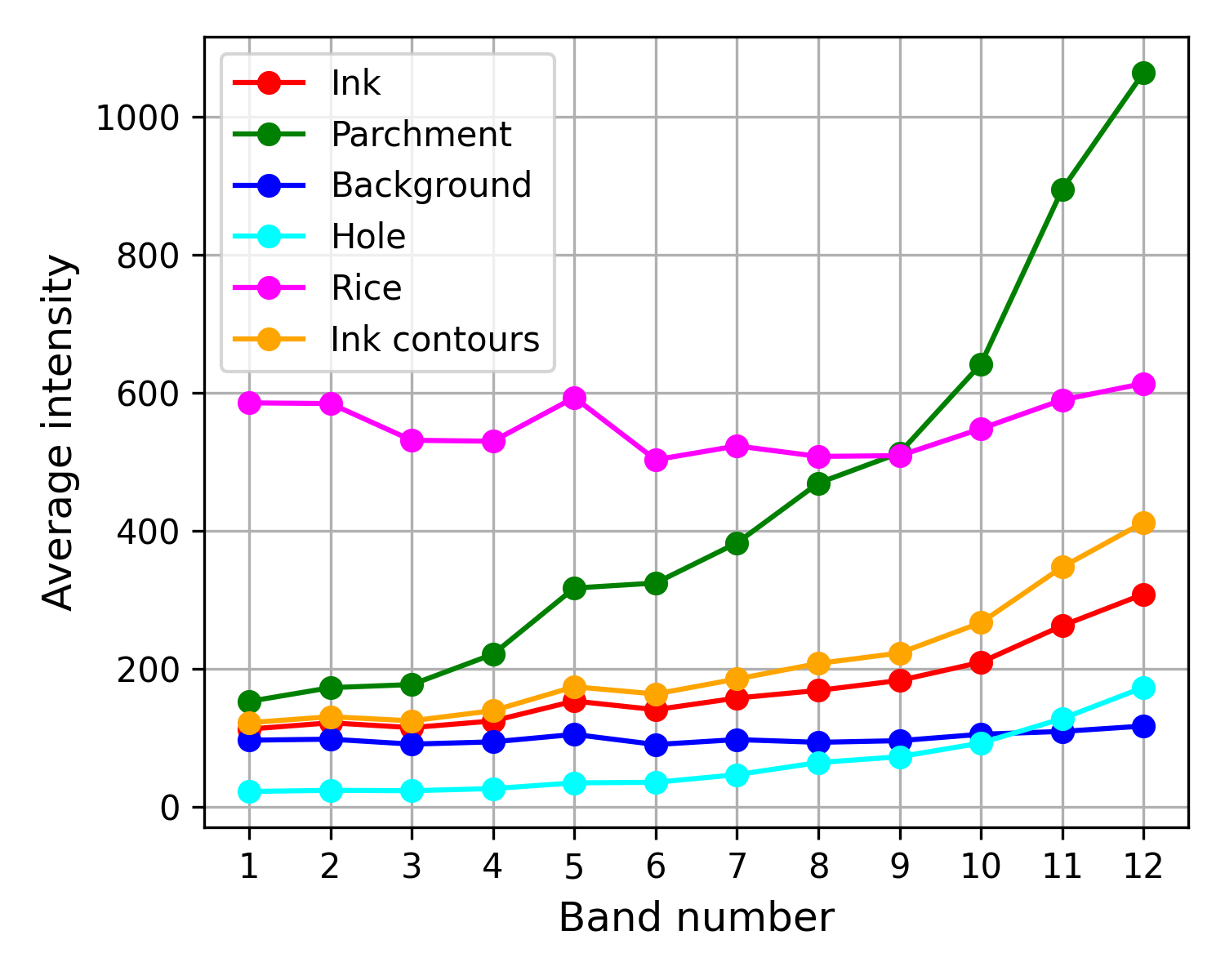}
  \caption{Average intensity trends showing that ink contours are more distinguishable than ink regions due to their thinner ink layer, allowing greater influence from the underlying parchment}
  \label{fig:five_regions_with_ink_contour_ms_trends}
\end{figure}

An interesting observation is that ink contours may serve as the most reliable areas, preserving the shape information of the ink while minimizing internal ink region noise interference. Furthermore, the multispectral trend of ink contours differs significantly from that of holes and background due to their thinner ink layer, which allows greater influence from the underlying parchment. This makes ink contours more distinguishable than ink pixels alone (Figure~\ref{fig:five_regions_with_ink_contour_ms_trends}).

\begin{figure}[htb]
  \centering
  \includegraphics[width=.95\linewidth]{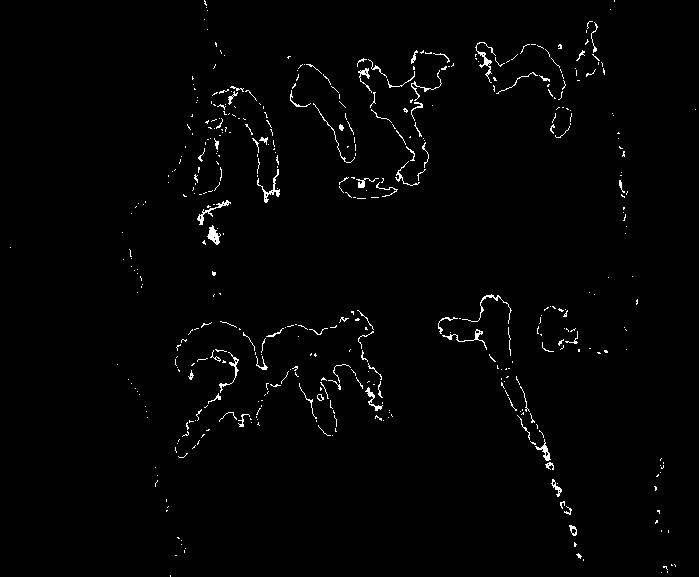}
  \caption{Multispectral thresholding result for ink contours using a percentile of \(n=10\)}
  \label{fig:ink_contour_10_zoom}
\end{figure}

For thresholding ink contours, we followed the same process as with ink, focusing only on the ink contours in the annotation of fragment 124-001. We applied the \(n\)-th percentile on both \(I_1\) and \(I_{12} - I_1\), selecting \(n=10\), which effectively highlighted the ink contours (Figure~\ref{fig:ink_contour_10_zoom}).

\subsection{Energy Minimization}

In this section, we formulate an energy minimization approach to enhance the segmentation of ink and parchment regions in binary document images. We begin by defining an energy function for the segmentation task, balancing the cost of assigning pixels to certain labels with the desire for smoothness between neighboring pixels. We then show how this general energy minimization framework is specifically adapted to our problem. By incorporating initial masks obtained from multispectral thresholding, we tailor the energy function to refine the ink contour masks and improve the delineation between ink, parchment, and other regions.

\subsubsection{Energy minimization formulation}
Let $\mathcal{L}$ be the set of two regions: foreground and background, and let $\mathcal{P}$ be the set of pixels in the binary image to be segmented. Energy minimization finds a labeling $f$ that assigns each pixel $p \in \mathcal{P}$ to a label $\ell_p \in \mathcal{L}$, such that the energy function $\textbf{E}(f)$ is minimized:
\[
\textbf{E}(f) = \sum_{p \in \mathcal{P}} D(p, \ell_p) + \sum_{\{p,p'\} \in \mathcal{N}} d(p, p') \cdot \delta(\ell_p \neq \ell_{p'})
\]
Here, $D$ represents the data cost, $d$ represents the smoothness cost, and $\delta$ is an indicator function. The data cost, $D(p, \ell_p)$, measures the cost of assigning pixel $p$ to label $\ell_p$, which is defined using the Euclidean distance between pixel $p$ and the nearest region in $\mathcal{L}$. The smoothness cost, $d(p, p')$, penalizes the assignment of different labels to neighboring pixels, encouraging similar labels for neighboring pixels. Let $\mathcal{N}$ be the set of nearest neighboring pixel pairs. Then, for all $\{p, p'\} \in \mathcal{N}$:
\[
d(p, p') = 
\begin{cases}
1, & \text{if } p \text{ and } p' \text{ are adjacent white pixels} \\
0, & \text{otherwise}
\end{cases}
\]
The function $\delta(\ell_p \neq \ell_{p'})$ equals $1$ if the condition inside the parentheses holds, and $0$ otherwise.

\subsubsection{EM Adaptation for segmenting ink and parchment}

\begin{figure*}[htb]
  \centering
  \includegraphics[width=.95\linewidth]{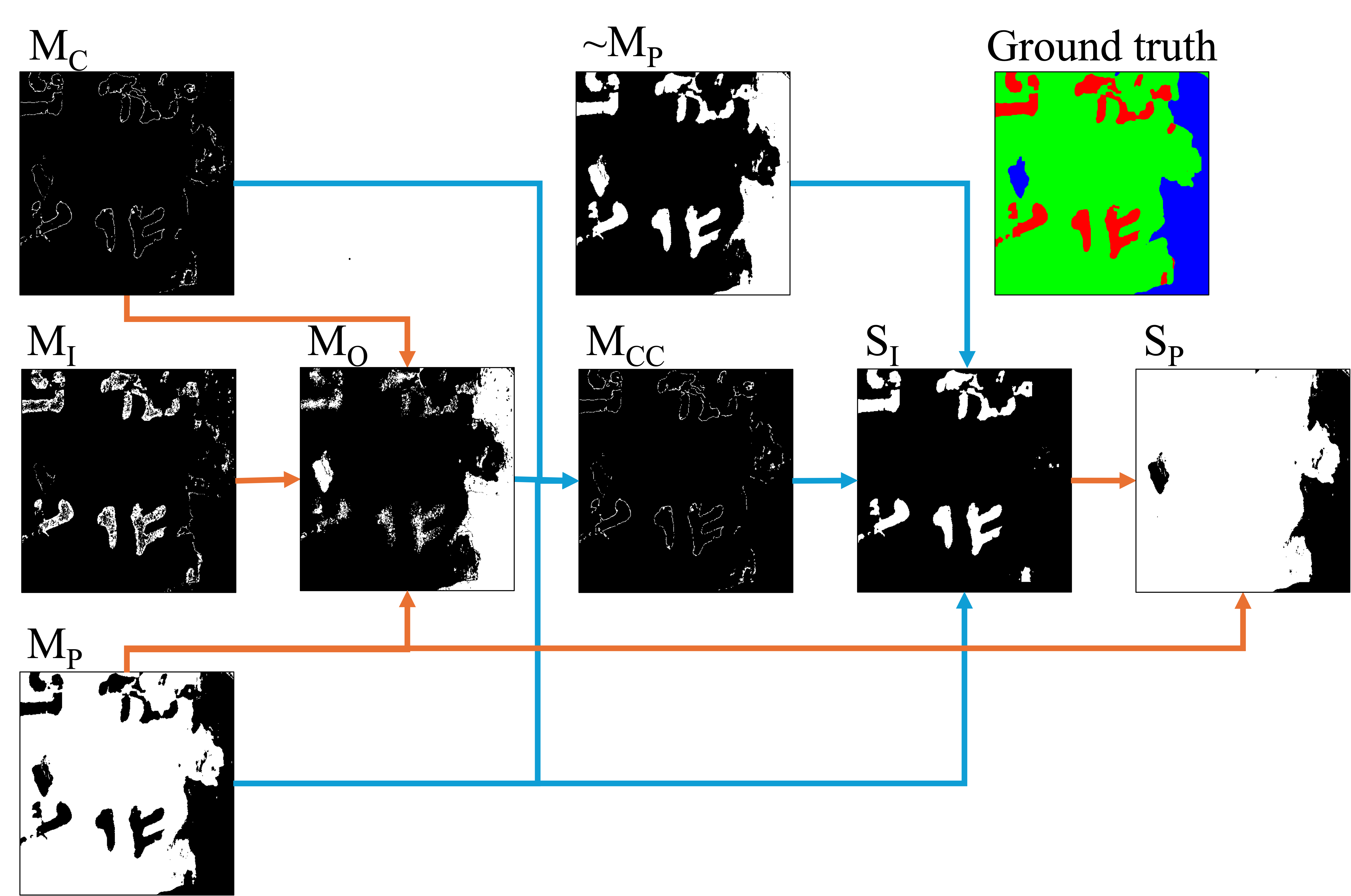}
  \caption{Image patches representing the inputs and outputs at different stages of the whole method pipeline. Red arrows indicate the union operation, while blue arrows denote the EM operation. The final outputs of the method are the ink segmentation ($S_I$) and parchment segmentation ($S_P$). The Ground Truth (GT) shows ink in red, parchment in green, and background in blue}
  \label{fig:method_pipeline}
\end{figure*}

As illustrated in Figure~\ref{fig:method_pipeline}, which shows the inputs and outputs at different stages of our method pipeline, we  begin with a parchment mask $M_P$, an ink mask $M_I$, and an ink contour mask $M_C$, all derived from multispectral thresholding (Section~\ref{sec:multispectral_thresholding}).
To obtain a mask for all other regions, $M_O$, we calculate the complement of the union of $M_P$, $M_I$, and $M_C$, defined as:
\[
M_O = (M_P \cup M_I \cup M_C)^\prime
\]

Next, we use energy minimization to refine the ink contour mask $M_C$ and obtain the clean ink contour mask, denoted as $M_{CC}$, which is derived as:
\[
M_{CC} = \arg \min_{f} E(f)
\]
where $f$ is the labeling function that assigns each pixel $p \in M_C$ to a label $\ell_p \in \mathcal{L}$, with $\mathcal{L} = \{ M_O, M_P \}$. This process stretches the mask $M_C$ towards the parchment region $M_P$, while balancing this movement by stretching the noise regions using $M_O$. This approach is chosen because the clean ink contours are expected to lie adjacent to the parchment, at the outermost boundaries of the ink regions, whereas noise is expected to lie adjacent to the background, hole, and rice regions.

After refining the ink contour mask, we further apply energy minimization to extract the ink segmentation, denoted as $S_I$, which is derived as:
\[
S_I = \arg \min_{f} E(f)
\]
where $f$ is the labeling function that assigns each pixel $p \in \neg M_P$ to a label $\ell_p \in \mathcal{L}$, with $\mathcal{L} = \{ M_P, M_{CC}\}$. This process stretches the complement of the parchment mask, $M_P^\prime$, which is assumed to contain all ink regions without internal noise, towards the refined ink contour mask $M_{CC}$, while using $M_P$ to counteract all non-ink regions. This ensures that all noise pixels fully enclosed by ink contours are included in $S_I$, and it delineates the boundary between the ink regions and the background, hole, and rice regions.

Finally, to obtain the parchment segmentation, denoted by $S_P$, we compute the union of the parchment mask $M_P$ and the ink segmentation $S_I$, resulting in:
\[
S_P = M_P \cup S_I
\]

\section{Experiments}

We conducted a series of experiments aimed at answering key questions and determining optimal parameter values for the MTEM method. For the evaluation of segmentation results, we used the metrics, Intersection over Union (IoU), F1-score, precision, and recall.

\subsection{Time efficiency}

\begin{figure}[htb]
  \centering
  \includegraphics[width=.95\linewidth]{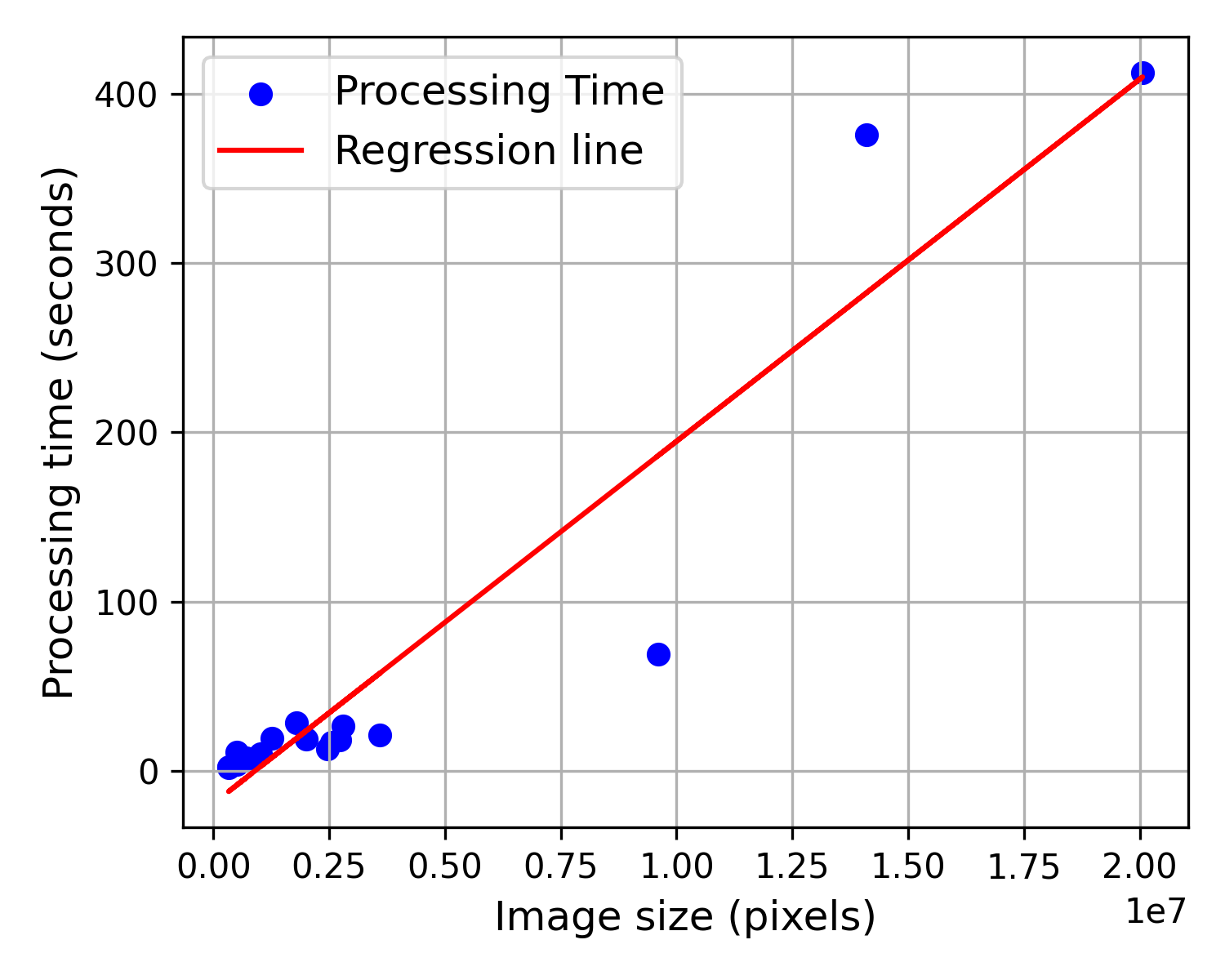}
  \caption{Relationship between image size and processing time for the MTEM segmentation method. The plot illustrates that processing time increases approximately linearly with the size of the input image}
  \label{fig:time_versus_image_size}
\end{figure}

We evaluated the time efficiency of the MTEM method by measuring execution time for each image in relation to its size. Figure~\ref{fig:time_versus_image_size} demonstrates that processing time increases with image size. We observed that larger images of approximately 20 million pixels, required over 300 seconds to process, while smaller images of approximately 500,000 pixels had processing times of only around 5 seconds. This suggests a roughly linear relationship between image size and processing time. The average processing time for an image from the QSD is 54 
seconds. These results highlight the scalability of the method, but also indicate the possible need for optimization with larger image sizes.

\subsection{Effectiveness of cleaning ink contours from noise}

We conducted experiments to assess the impact of cleaning the ink contours from noise by comparing two different segmentation results: one using noisy ink contours ($M_C$) and another using cleaned ink contours ($M_{CC}$). Ink contours exhibit a distinct multispectral trend compared to both ink regions and parchment regions. The trend of ink contour regions falls between that of ink and parchment, allowing them to serve as useful features for segmentation. However, noisy pixels with similar intensity values may be present in other regions, introducing errors. It is assumed that clean ink contours are positioned close to parchment regions, given that ink is present on the surface parchment. Therefore, we utilized $M_O$ to shrink the influence of the noise regions, and $M_P$ to expand the influence of the clean ink contours. Table~\ref{tab:clean_vs_noisy_results} presents the metrics obtained from both experiments. 

\begin{table*}[htb]
    \centering
    \caption{Comparison of segmentation results with noisy vs. cleaned ink contours.}
    \label{tab:clean_vs_noisy_results}
    \begin{tabular}{lrrrr}
        \hline
        \textbf{Method} & \textbf{IoU} & \textbf{Precision} & \textbf{Recall} & \textbf{F1-Score} \\
        \hline
        Ink (Noisy Contours) & 0.1990 & 0.2241 & 0.6953 & 0.2595 \\
        Ink (Clean Contours) & 0.6713 & 0.8935 & 0.7029 & 0.7676 \\
        Parchment (Noisy Contours) & 0.6355 & 0.6396 & 0.9906 & 0.7416 \\
        Parchment (Clean Contours) & 0.9764 & 0.9945 & 0.9818 & 0.9877 \\
        \hline
    \end{tabular}
\end{table*}

\begin{figure}[htb]
  \centering
  \includegraphics[width=.95\linewidth]{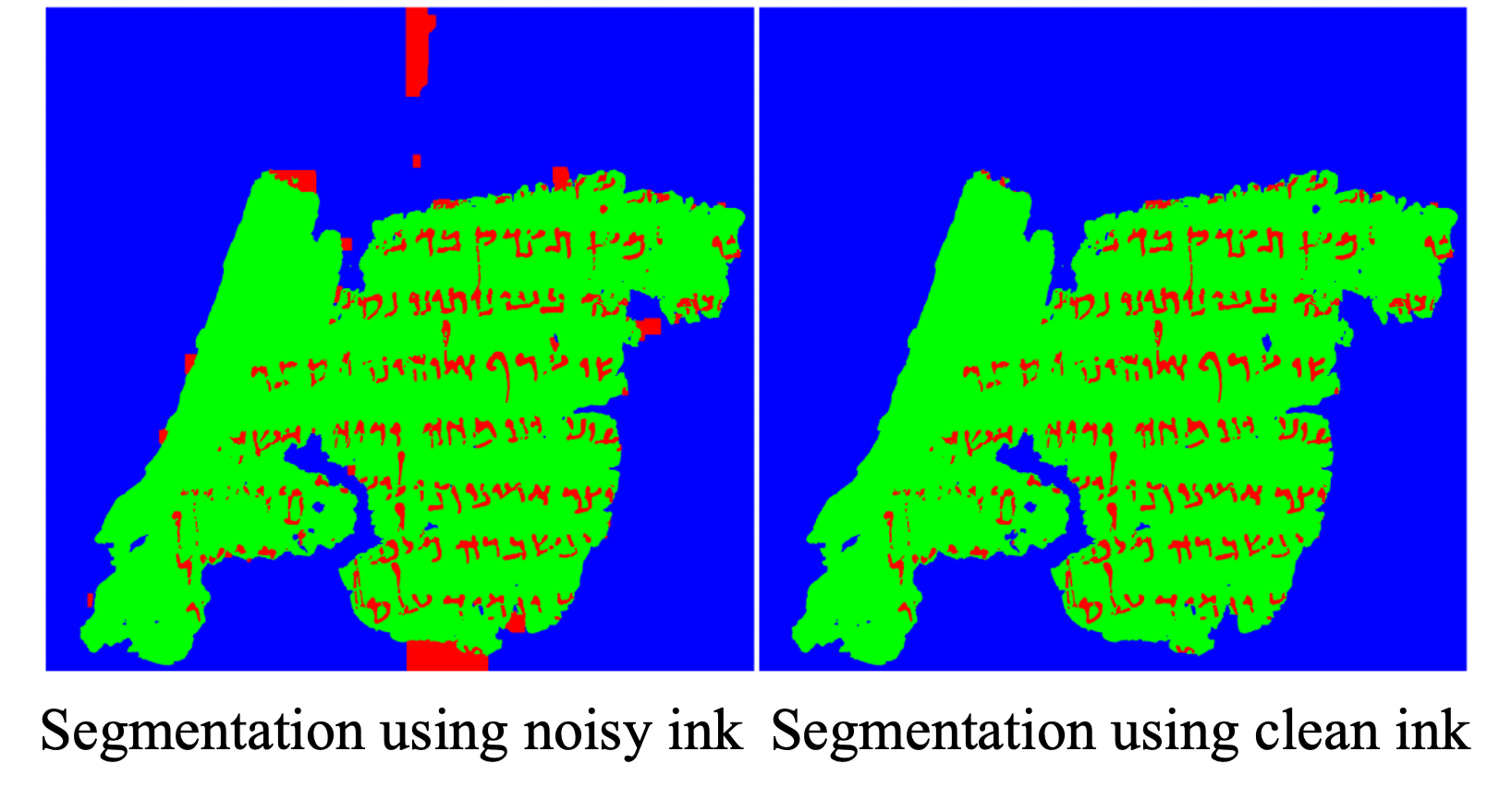}
  \caption{Comparison of segmentation results with noisy vs. cleaned ink contours}
  \label{fig:clean_vs_noisy_ink_segmentation}
\end{figure}

The results indicate that using cleaned ink contours improves ink segmentation ($S_I$) performance. The high recall and low precision values with noisy contours indicates that $S_I$ included more pixels from the other regions and $M_P$ remain unaffected but the final parchment segmentation, $S_P = M_P \cup S_I$, is influenced by the errors introduced from $S_I$ (Figure~\ref{fig:clean_vs_noisy_ink_segmentation}).

\subsection{Effectiveness of ink contour thickness}

Having determined that ink contours are crucial for the performance of the MTEM method, an important question arises: how thick should these contours be to achieve optimal performance? We extracted ink contours with varying thickness values from the annotations of fragment 124-001, which were used to compute the $n=10$ percentile values for \(I_1\) and \(I_{12} - I_1\). We then evaluated the impact of these different thickness values on the segmentation results for ink and parchment regions.

\begin{figure}[htb]
  \centering
  \includegraphics[width=.95\linewidth]{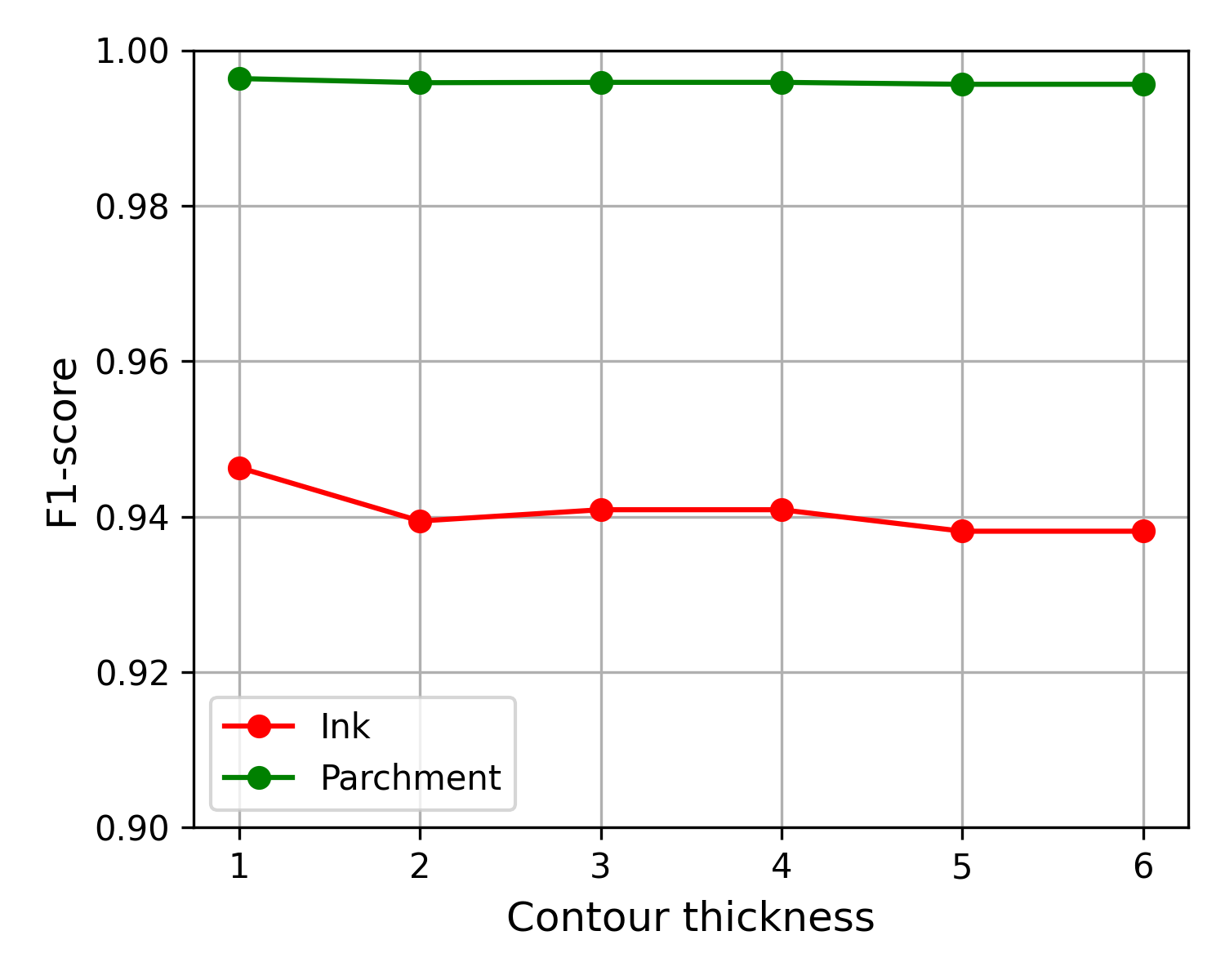}
  \caption{Effect of contour thickness on performance}
  \label{fig:performance_vs_contour_thickness}
\end{figure}

\begin{figure}[htb]
  \centering
  \includegraphics[width=.95\linewidth]{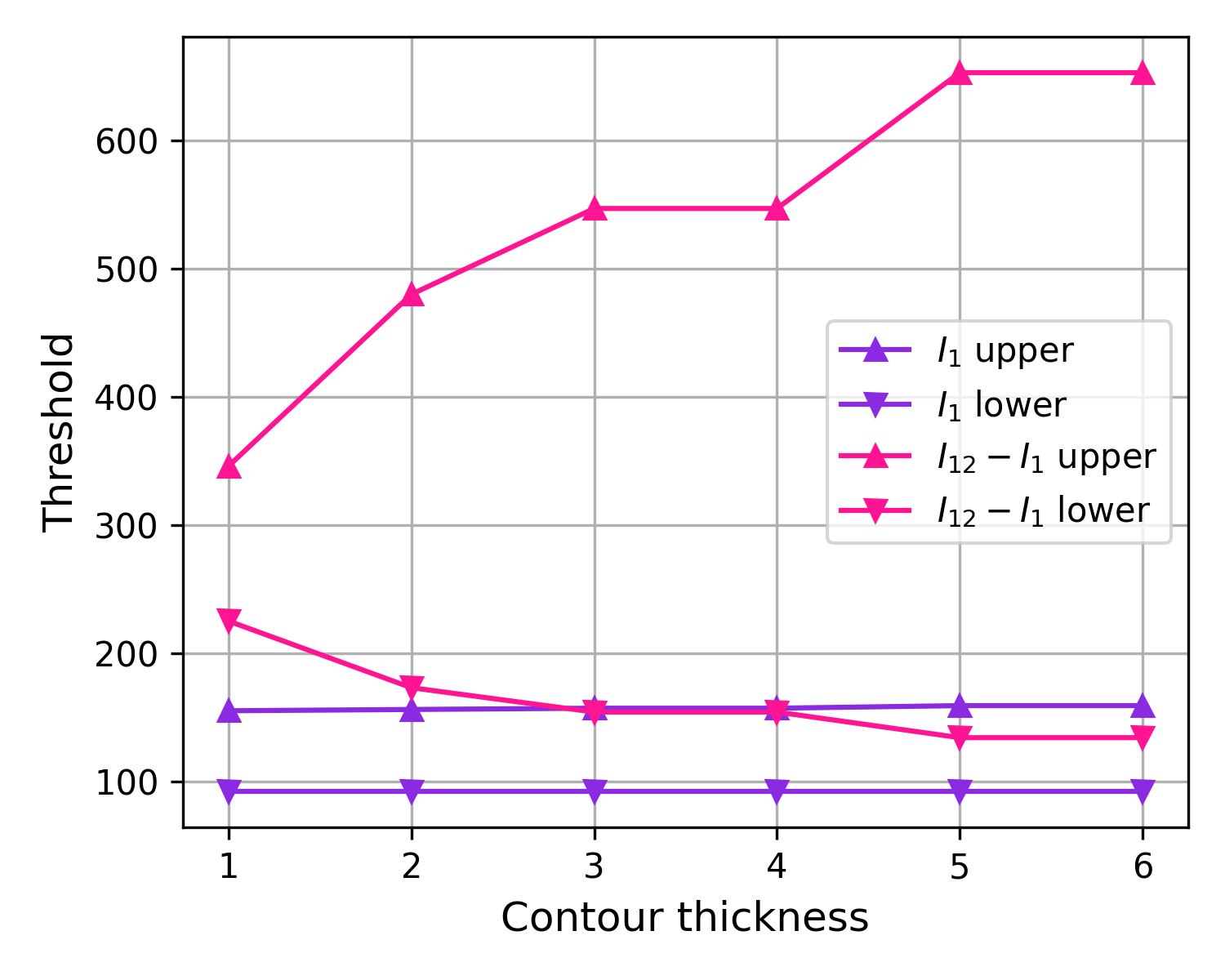}
  \caption{Effect of contour thickness on threshold values}
  \label{fig:contour_thresholds_vs_contour_thickness}
\end{figure}

We observed that contour thickness plays a significant role in distinguishing ink from other regions, such as rice, hole, and background areas because, while the accuracy of ink segmentation ($S_I$) decreases, the performance of parchment segmentation ($S_P$) remains stable as contour thickness increases (Figure~\ref{fig:performance_vs_contour_thickness}). The increasing upper threshold and decreasing lower threshold for \(I_{12} - I_1\) as contour thickness increases, indicate that contours extend into both the parchment region and ink region (Figure~\ref{fig:contour_thresholds_vs_contour_thickness}). Additionally, the constant \(I_1\) threshold across varying contour thicknesses confirms that \(I_1\) alone is insufficient for discriminating between ink and parchment pixels, as illustrated in Figure~\ref{fig:band12_minus_band1_versus_band_1_intensity_distribution}.


\subsection{Optimal percentile}\label{sec:optimal_percentile}

\begin{figure}[htb]
  \centering
  \includegraphics[width=.95\linewidth]{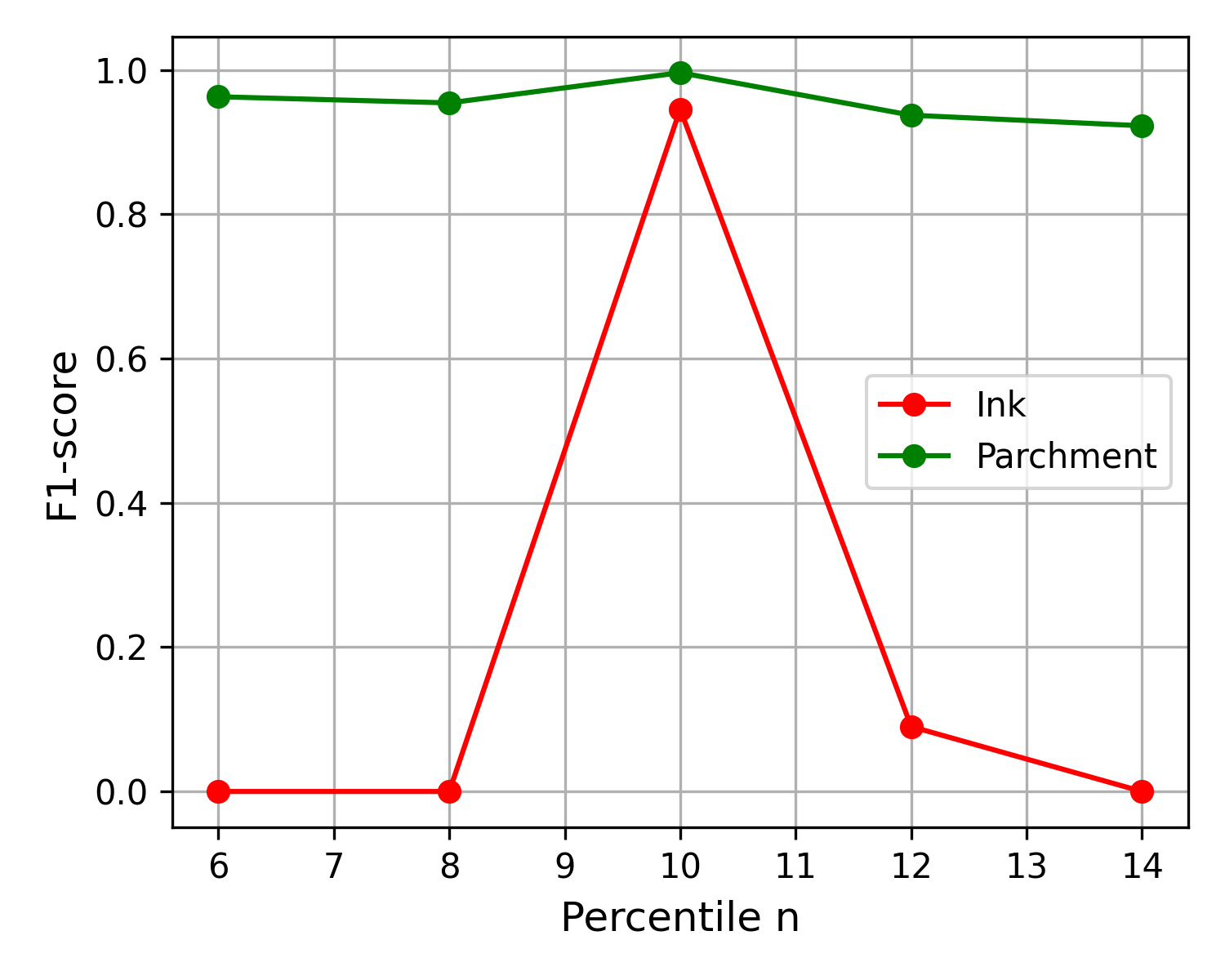}
  \caption{Effect of percentile value $n$ on the segmentation performance for ink and parchment. The figure illustrates that $n = 10$ yields the highest F1-score, especially for ink segmentation}
  \label{fig:ablation_for_percentiles}
\end{figure}

We explore the optimal percentile for the performance of the MTEM method in ink and parchment segmentation. The results (Figure~\ref{fig:ablation_for_percentiles}) for ink segmentation indicate that the choice of percentile has a significant impact on performance. At all tested values of n, the ink segmentation F1-score was almost zero, except when $n = 10$. For parchment segmentation, $n = 10$ also proved to be the optimal value, although its trend across all n values remained relatively consistent. This is because the separability bounds for ink are very close to those of other regions, while the separability bounds for parchment are much more distinct from the other regions, as illustrated in Figure~\ref{fig:band12_minus_band1_versus_band_1_intensity_distribution}. Overall, the results suggest that the optimal percentile value is $n = 10$ and highlights the need for careful selection of percentile thresholds to ensure robust performance.

\section{Results}

\subsection{Quantitative results}
In this section, we present the evaluation of various ink and parchment segmentation methods applied to QSD images. The evaluated methods include the proposed MTEM method for both ink and parchment segmentation, along with traditional binarization techniques like Otsu and Sauvola for parchment segmentation. The evaluation metrics used to assess segmentation performance are Intersection over Union (IoU), precision, recall, and F1-Score. The results are summarized in Table~\ref{tab:segmentation_results}.

\begin{table*}[htb]
\caption{Performance of different methods applied to QSD images for segmenting ink and parchment regions.}
    \centering
    \begin{tabular}{lrrrr}
        \hline
        \textbf{Method} & \textbf{IoU}  & \textbf{Precision} & \textbf{Recall} & \textbf{F1-Score} \\
        \hline
        MTEM ink & 0.6713  & 0.8935 & 0.7029 & 0.7676 \\
        MTEM parchment & 0.9764  & 0.9945 & 0.9818 & 0.9877 \\
        Otsu parchment & 0.5642  & 0.5892 & 0.9460 & 0.6919 \\
        Sauvola parchment & 0.3104  & 0.3192 & 0.9293 & 0.4624 \\
        Otsu + Sauvola parchment & 0.5594 & 0.5924 & 0.9277 & 0.6888 \\
        \hline
    \end{tabular}   
    \label{tab:segmentation_results}
\end{table*}

The proposed MTEM method for ink segmentation yields promising results, with an IoU of 0.6713. The precision is 0.8935, indicating that most of the detected ink pixels are indeed correct. However, the recall is 0.7029, suggesting that there is still room for improvement in capturing all ink pixels, as some are missed during detection. We observe that these missed pixels often occur in ink regions contaminated by unspecified substances (such as salt) or in ink areas on dark parchment surfaces, making their contours less detectable.

For parchment segmentation, the MTEM method performs exceptionally well, achieving an IoU of 0.9764. The high F1-Score of 0.9877 further demonstrates the reliability of MTEM for parchment segmentation.

In contrast, the performance of the Otsu method, the Sauvola method, and their combination (via an AND operation) for parchment segmentation is notably lower compared to MTEM. The relatively low precision and high recall indicate that these methods tend to over-segment, capturing almost all potential parchment regions but also producing a significant number of false positives, especially in rice regions that have high reflectivity in the infrared images.

Overall, the results indicate that the MTEM method outperforms traditional binarization techniques such as Otsu and Sauvola for parchment segmentation. Specifically, MTEM demonstrates significantly higher IoU and F1-Score, highlighting its effectiveness in detecting parchment areas. Additionally, the MTEM method segments ink regions and can distinguish ink areas from holes and background regions.

\subsection{Qualitative results}

\begin{figure*}[ht]
  \centering
  \includegraphics[width=.95\linewidth]{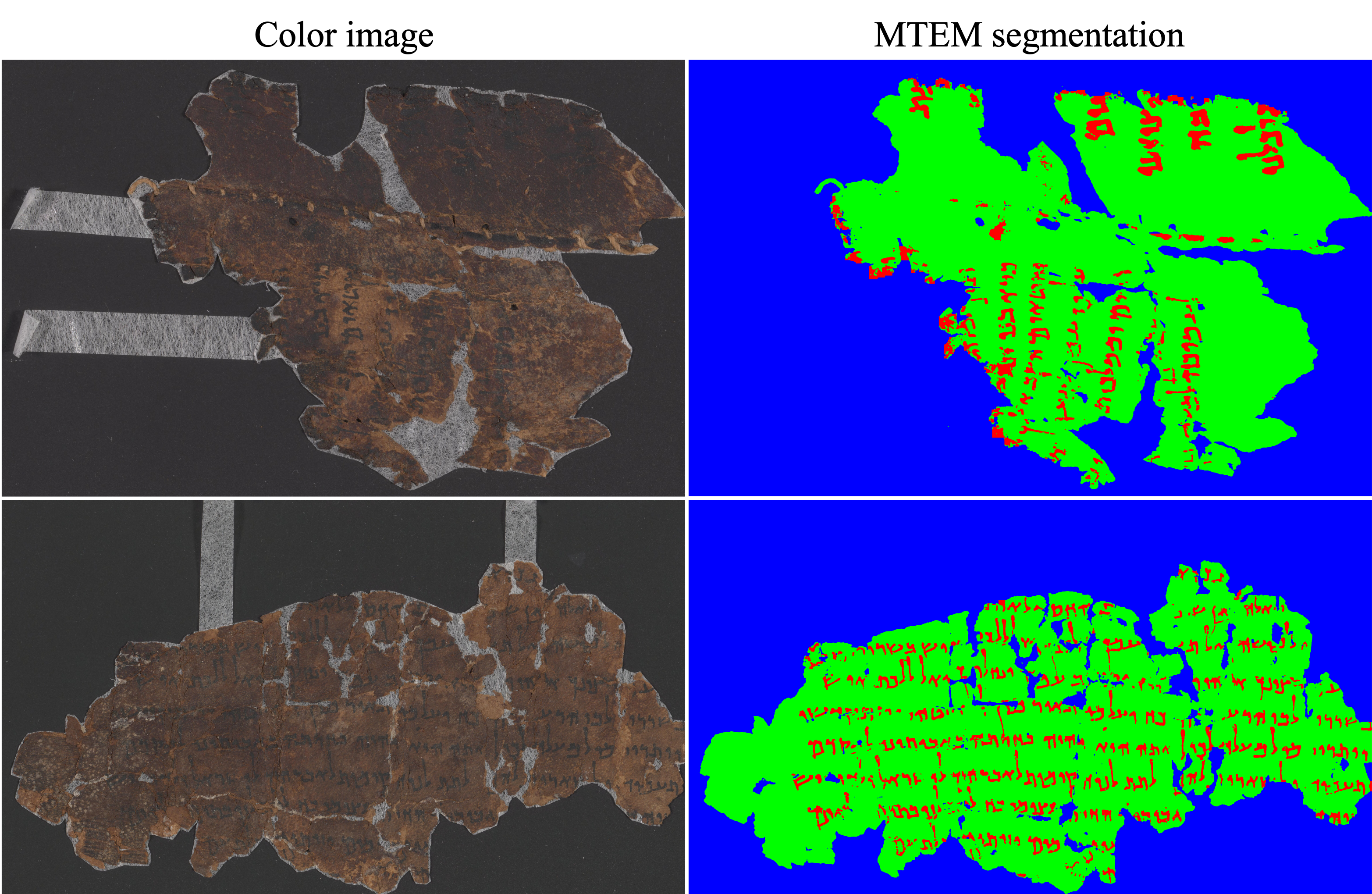}
  \caption{Qualitative results demonstrating the effectiveness of the MTEM segmentation method. The left column shows the QSD  color images, while the right column displays their segmentation using MTEM. In the segmentation images, ink is represented in red, parchment in green, and background in blue. The MTEM method can discriminate ink from holes and background, removes rice paper regions even when they are between the parchment and ink regions, and accurately delineates the edges of ink regions that are adjacent to the background or rice regions.}
  \label{fig:qualitative_result}
\end{figure*}

To illustrate the effectiveness of the MTEM segmentation method, we present some QSD images and their segmented outputs using MTEM in Figure~(\ref{fig:qualitative_result}). Most importantly, MTEM can discriminate ink from holes and background, remove rice paper regions even when they are between the parchment and ink regions, and accurately delineate the edges of ink regions that are adjacent to the background or rice regions.

The majority of error cases occur when ink is present on a very dark parchment region. In such instances, the MTEM method apparently fails to extract the ink contours and consequently fails to segment the ink. A second type of error case we observed occurs on ink regions influenced by an unknown but apparently white substance, which prevents the ink contours from being extracted and, as a result, the ink region cannot be segmented.

\begin{figure*}[ht]
  \centering
  \includegraphics[width=.95\linewidth]{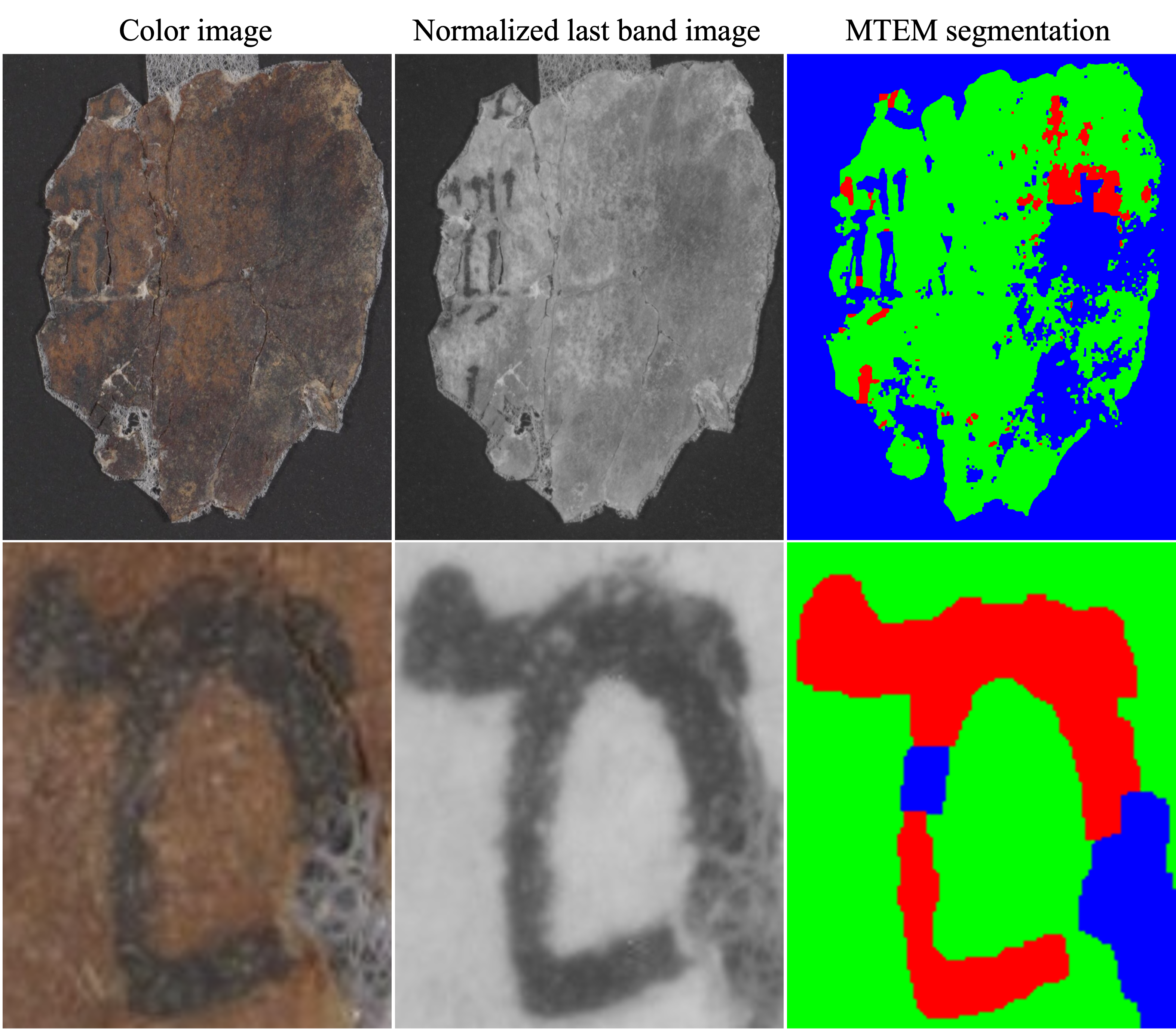}
  \caption{Illustration of error cases in MTEM segmentation. The left column shows the QSD color images, the middle column displays the normalized last band images, and the right column presents the MTEM segmentation results. In the first example, ink is present on a very dark parchment region, causing the MTEM method to fail in extracting the ink contours and consequently fail to segment the ink. In the second example, the ink regions are influenced by an unknown but apparently white substance, which prevents the ink contours from being extracted, and as a result, the ink regions cannot be segmented.}
  \label{fig:error_cases}
\end{figure*}

\section{Conclusion}

This study presents a novel approach for segmenting ink and parchment in Dead Sea Scroll fragments using multispectral thresholding and energy minimization techniques. The proposed Multispectral Thresholding and Energy Minimization (MTEM) method addresses the complex challenges of separating ink from the black background and eliminating noise within ink regions. MTEM demonstrates significant improvements over traditional binarization methods, effectively isolating ink and parchment regions even under varied conditions of degraded manuscripts. This adaptability suggests its potential for application to other multispectral image datasets, such as the MultiSpectral Document Binarization (MSBIN) dataset~\cite{hollaus2019cnn}, the MultiSpectral Text Extraction (MS-TEx) dataset~\cite{hedjam2015icdar}, and the Multispectral Hebrew Ostraca dataset~\cite{faigenbaum2012multispectral}.

Future work will extend these segmentations to improve Kraken’s~\cite{kiessling2019kraken} character identification and correct inaccurate detections, enabling precise character segmentations. By integrating segmented characters and parchments, further analyses of DSS images will be facilitated, including the alignment of transcriptions with images to add a searchable textual layer to the IAA images, allowing scholars to locate text in its spatial context.

\end{document}